\documentclass[letterpaper,journal]{IEEEtran}
\usepackage[caption=false,font=normalsize,labelfont=sf,textfont=sf]{subfig}
\usepackage{capt-of}
\usepackage{multicol}
\usepackage[bookmarks=true,hidelinks]{hyperref}
\usepackage{graphicx}
\usepackage{amsmath}
\usepackage{amssymb}
\usepackage{booktabs}
\usepackage{multirow}
\usepackage[table,dvipsnames]{xcolor}
\usepackage{stfloats}
\usepackage{placeins}
\usepackage{textcomp}
\usepackage{stfloats}
\usepackage{url}
\usepackage{verbatim}
\usepackage{graphicx}
\usepackage{cite}
\usepackage{balance}
\usepackage{enumitem}

\usepackage[mode=match]{siunitx}
\DeclareSIUnit\pixel{px}

\newcommand{\W}{\mathtt{W}}
\newcommand{\C}{\mathtt{C}}
\newcommand{\A}{\mathtt{A}}
\newcommand{\nmat}[1]{\mathbf{#1}}
\newcommand{\pose}[2]{\nmat{T}_{#1 #2}}
\newcommand{\nvec}[1]{\boldsymbol{#1}}
\newcommand{\point}[2]{\nvec{#1}_{#2}}

\definecolor{hm0}{HTML}{FFFFFF}
\definecolor{hm1}{HTML}{FFF5CC}
\definecolor{hm2}{HTML}{FFE6AA}
\definecolor{hm3}{HTML}{FFD699}
\definecolor{hm4}{HTML}{FFC499}
\definecolor{hm5}{HTML}{FFB299}
\definecolor{hm6}{HTML}{FF9F99}
\definecolor{hm7}{HTML}{FF8C8C}
\definecolor{headerblue}{HTML}{7F8FA6}

\def\secref#1{Sec.~\ref{#1}}
\def\appref#1{Appendix ~\ref{#1}}
\def\figref#1{Fig.~\ref{#1}}

\def\eqref#1{Eq.~(\ref{#1})}

\hypersetup{
   pdftitle={Dense Endoluminal Mapping and Localisation from Monocular Bronchoscopy},
   pdfauthor={Salvatore Esposito, Matias Mattamala, Daniel Rebain, Liang Zhao, Kev Dhaliwal, Mohsen Khadem, Subramanian Ramamoorthy},
   pdfsubject={Robotics},
   pdfkeywords={SLAM, Localization, Mapping, Deep Learning for Visual Perception, Computer Vision for Medical Robotics, Medical Robots and Systems, Factor Graphs, Bronchoscopy}
}

\begin{document}

\title{Dense Endoluminal Mapping and Localisation\\ from Monocular Bronchoscopy}

\author{Salvatore Esposito$^{1}$, Matias Mattamala$^{1}$, Daniel Rebain$^{2}$, Liang Zhao$^{1}$, Kev Dhaliwal$^{1}$, Mohsen Khadem$^{1}$, Subramanian Ramamoorthy$^{1}$\\
$^{1}$University of Edinburgh \quad $^{2}$University of British Columbia}
\maketitle

\begin{abstract}
Bronchoscopy inserts a camera through the airways to diagnose endoluminal disease. Locating the camera along the airway usually requires additional equipment, such as an electromagnetic tracker and a preoperative CT scan.
In this paper, we investigate the use of the monocular camera stream to localise the bronchoscope without requiring additional sensing. We introduce a monocular SLAM system that localises the tip of a bronchoscope and reconstructs the airway surface around it from RGB video alone. A Depth Anything 3 backbone, fine-tuned on over a million photorealistic airway renderings, returns metric per-pixel depth and is the system's only source of scale. DINOv3 features routed through a custom cross-attention matcher supply sub-pixel correspondences on mucosa where classical descriptors fail. Both enter a factor graph alongside a dense constraint from projective iterative closest point (ICP) registration against the system's own running reconstruction, so surface evidence stabilises the directions the visual factors leave weakly observable.
On six airways rendered from patient CT scans, the system reaches \SI{26.1}{\milli\meter} mean absolute trajectory error and \SI{77.5}{\percent} mean surface F-score. Its mean surface accuracy exceeds that of five recent monocular SLAM and feed-forward baselines. MASt3R-SLAM, which fits a free scale to the ground truth, has a lower mean trajectory error (\SI{22.0}{\milli\meter}). ORB-SLAM3 completes only two of the six airways, whereas our system completes all six and also reconstructs two ex vivo specimens.
\end{abstract}


\section{Introduction}
\label{sec:introduction}
\IEEEPARstart{B}{ronchoscopy} is a procedure for examining and sampling the airway~\cite{durand2013british}. Asthma, chronic obstructive pulmonary disease, cystic fibrosis and endoluminal tumours affect the airway walls or obstruct the lumen~\cite{james2007remodelling, hogg2004pathophysiology, boucher2007cf, ernst2004central}. In each case an operator advances a bronchoscope to a specific airway and works under direct vision to sample or inspect it.

\begin{figure}[!t]
  \centering
  \includegraphics[width=\linewidth]{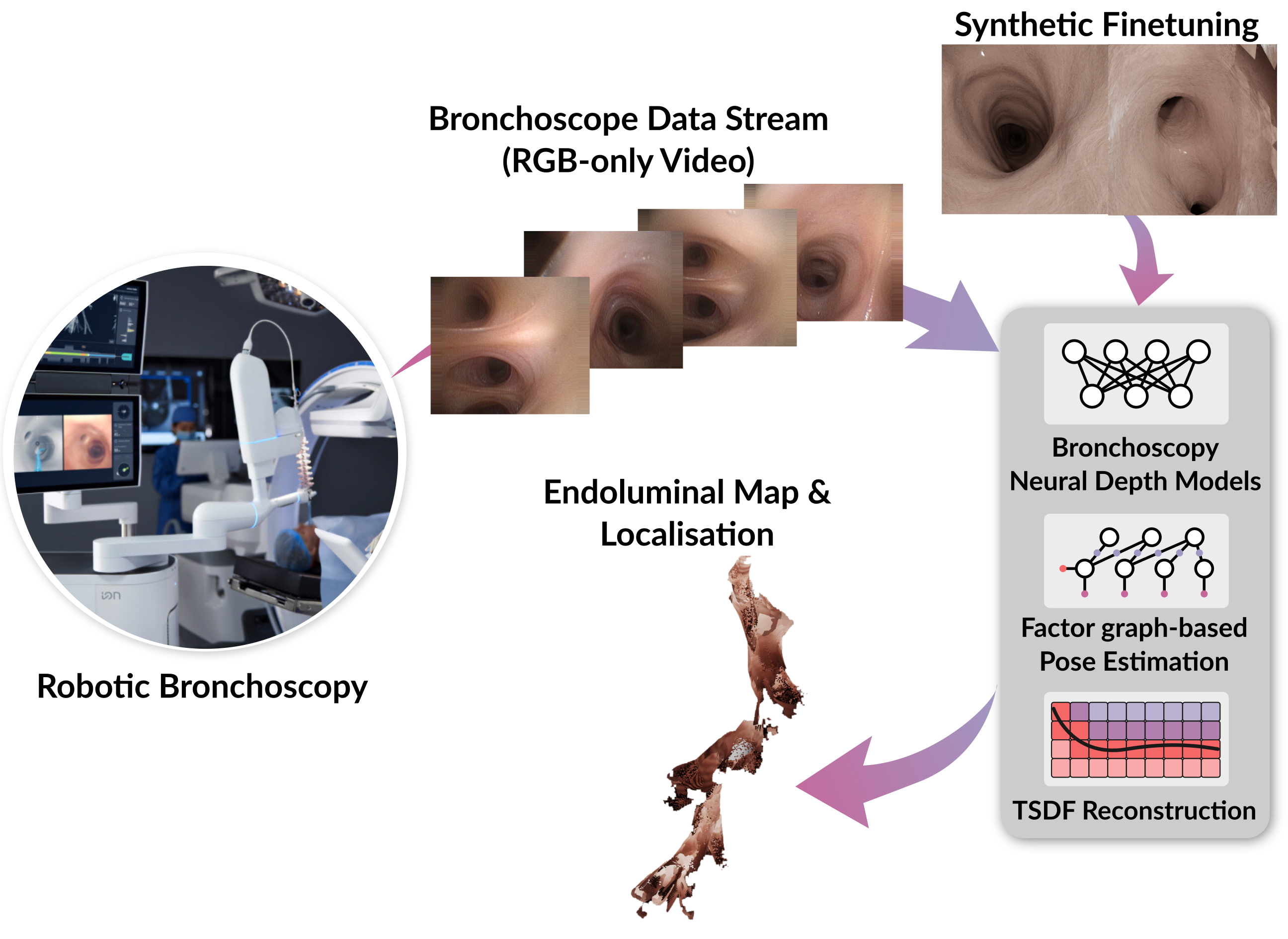}
  \caption{\textbf{From bronchoscope video to an endoluminal map.} A robotic bronchoscope (left) streams RGB video of the airway interior. This video feds our system, which estimates the camera pose and reconstructs the airway wall, using learnt modules trained on synthetic bronchoscopy data (top right). The robot image is taken from the Ion system page of Intuitive~\cite{intuitiveion}.}
  \label{fig:teaser}
\end{figure}

In current practice the tip is located with equipment brought in alongside the bronchoscope: an electromagnetic field generator tracks the catheter, and that tracking is registered to a computed tomography (CT) airway map acquired before the procedure~\cite{gildea2006emn}. This locates the catheter but recovers nothing about the surface it observes, and its accuracy is bounded by the CT-to-patient registration and by repeat CT imaging that accumulates ionising radiation~\cite{brenner2007computed}.

The camera at the tip of the bronchoscope passes within millimetres of the tissue. Some robotic bronchoscopes support navigation~\cite{zhang2024aicopilot} and camera-pose estimation~\cite{deng2023featurebased}, without mapping the surface. Dense reconstruction has been demonstrated mainly in the gastrointestinal tract~\cite{mahmoud2019live, morlana2024colonslam}. In the lung, the airway reconstruction we are aware of anchors its geometry to a preoperative CT scan~\cite{dunnbeltran2026stopholding}.

Four properties of endoluminal imaging stand between the video and that goal. First, the mucosa lacks enough texture to enable feature-based tracking, so endoscopic adaptations of established visual SLAM systems such as ORB-SLAM produce sparse maps and cannot hold one track from the start of a sequence to its end~\cite{grasa2014visual, mahmoud2019live, morlana2024colonslam}. Second, a single camera recovers geometry only up to an unknown factor, so neither the trajectory nor the map built by monocular SLAM reflects the real dimensions of the lumen. Third, the light source travels with the camera, so specular highlights slide across the tissue instead of staying fixed to it, and the depth estimated at those pixels is corrupted~\cite{mountney2010tissue}. Lastly, the airway is locally a tube, which leaves rotation about its axis and translation along it weakly observable, similar to the problem LiDAR SLAM faces in tunnels~\cite{zhang2016degeneracy}.

In this work, we introduce a monocular mapping and localisation system that addresses these four challenges, shown in Fig.~\ref{fig:teaser}. The system represents the airway wall as a truncated signed distance field (TSDF), a voxel grid in which each voxel stores its signed distance to the nearest surface, clipped to a narrow band around the surface. Our contributions are as follows.
\begin{enumerate}[leftmargin=*]
    \item \textbf{Bronchoscopy metric depth predictor.} A fine-tuned depth estimator that supplies the system's metric scale and per-pixel uncertainty. 
    \item \textbf{Dense matcher for textureless mucosa.} We propose a custom feature matcher based on a cross-attention head trained for sub-pixel matching of DINOv3 features within the airway.
    \item \textbf{Two-pass factor graph fusing visual measurements with a dense surface constraint.} We propose a factor graph formulation that combines visual landmark measurements and motion models, which first estimates an initial pose for TSDF-based ICP registration, and then refines the trajectory estimate with the ICP correspondences.
    \item \textbf{An experimental validation} on six airways rendered from patient CT scans, all from the public ATM'22 dataset~\cite{zhang2023atm22}, as well as ex vivo specimens, reaching $26.13$\,mm mean absolute trajectory error (ATE) and $77.47\,\%$ mean F-score in reconstruction. The mean surface accuracy exceeds that of five classical and feed-forward baselines. MASt3R-SLAM, which fits a free scale, has a lower mean ATE ($22.02$\,mm).
\end{enumerate}

\section{Related Work}
\label{sec:related}

\subsection{Medical Visual SLAM and Reconstruction}
Medical visual SLAM is more visually challenging than outdoor SLAM because the tissue lacks the texture ordinary scenes provide. For example,  ORB-SLAM~\cite{campos2021orbslam3} relies on extracted keypoints from corners; LSD-SLAM~\cite{engel2014lsd} and DSO~\cite{engel2018direct} align images by their pixel intensities instead, relying on a constant brightness assumption. The mucosa breaks these assumptions, which has been observed when transferring these methods to endoscopic applications, shown by a reduced number of matches and sparser maps~\cite{grasa2014visual}. For this reason, prior works have presented colonoscopy-specific approaches, such as handheld monocular reconstruction~\cite{mahmoud2019live}, multi-map colonoscopy SLAM~\cite{morlana2024colonslam} to target the gastrointestinal tract, or self-supervised depth predictors on endoscopic video~\cite{liu2020dense}.

Other endoscopic systems estimate tissue deformation together with the camera. DefSLAM~\cite{lamarca2021defslam} and NR-SLAM~\cite{gomezrodriguez2024nrslam} recover a deforming surface from monocular video, MIS-SLAM~\cite{song2018misslam} couples ORB-SLAM with dense deformable tracking on stereoscope video, and Zhang et al. carry the same treatment into endovascular intervention~\cite{zhang2023aortic}. These methods model deforming tissue. The airways show little deformation in our ex vivo ventilated-lung video, so we assume they are rigid and do not compare against these methods (Sec.~\ref{sec:limitations}).

Bronchoscopy research itself prioritises navigation over reconstruction. Catheter localisation against a preoperative CT airway map is the goal of bronchoscopy registration, electromagnetic tracking, and learned shared-control assistants such as the AI co-pilot bronchoscope robot~\cite{zhang2024aicopilot}. These recover the pose of the scope relative to the scan, and none reconstructs the lumen surface. Concurrent work fits dense Gaussian-splat reconstructions of bronchoscopy by coupling them to a paired inhale/exhale preoperative CT scan~\cite{dunnbeltran2026stopholding}, but still relies on the volumetric prior to anchor the geometry. The closest precedent is Deng et al.~\cite{deng2023featurebased}, who benchmarked feature-based visual odometry on bronchoscopy phantoms and ex vivo lungs against electromagnetic-tracking ground truth but did not produce a surface map. Of the systems above, each either anchors its geometry to a preoperative scan or stops at the catheter pose without reconstructing the lumen.

\subsection{Learning and Hybrid Approaches to 3D Reconstruction}
Such a system now needs measurements available from a single image. DINOv3~\cite{simeoni2025dinov3} describes every image patch with a learned feature vector, and those vectors stay distinctive on surfaces where ORB and SIFT find no corners to describe. Depth Anything 3~\cite{lin2025depthanything3} predicts depth, confidence and ray map from a single image, where stereo and structure from motion need several views to triangulate it. VGGT~\cite{wang2025vggt} predicts the geometry itself, returning camera parameters, depth maps and point maps for a set of images in one pass and without any geometric optimisation. DROID-SLAM~\cite{teed2021droid} also reconstructs with an end-to-end model, but keeps an explicit optimisation and learns the update it applies to optical flow, depth and pose.

Several SLAM systems build on these feed-forward models. MASt3R-SLAM~\cite{murai2025mast3rslam} tracks against the per-pixel 3D points a two-view model predicts and fuses them in real time, VGGT-SLAM 2.0~\cite{maggio2026vggtslam2} aligns per-chunk reconstructions on the SL(4) manifold, and ViSTA-SLAM~\cite{zhang2025vistaslam} pairs a two-view front end with a pose graph. Endo3R~\cite{guo2025endo3r} adapts the same idea to surgical video, predicting globally aligned point maps and camera poses from monocular endoscopy. In each, the learned model predicts the geometry and the back end combines the predictions.

Building upon these methods, we use learnt components in the front end to mitigate the four challenges described in the Introduction (\secref{sec:introduction}, which are trained on bronchoscopy-oriented data. However, we still rely on a factor graph formulation to fuse the different geometrical cues and their uncertainties, making our system hybrid.

\section{System Overview}
\label{sec:method}

\begin{figure*}[ht]
  \centering
  \captionsetup{font=footnotesize}
  \includegraphics[width=1.0\linewidth, trim=0pt 0pt 0pt 0pt, clip]{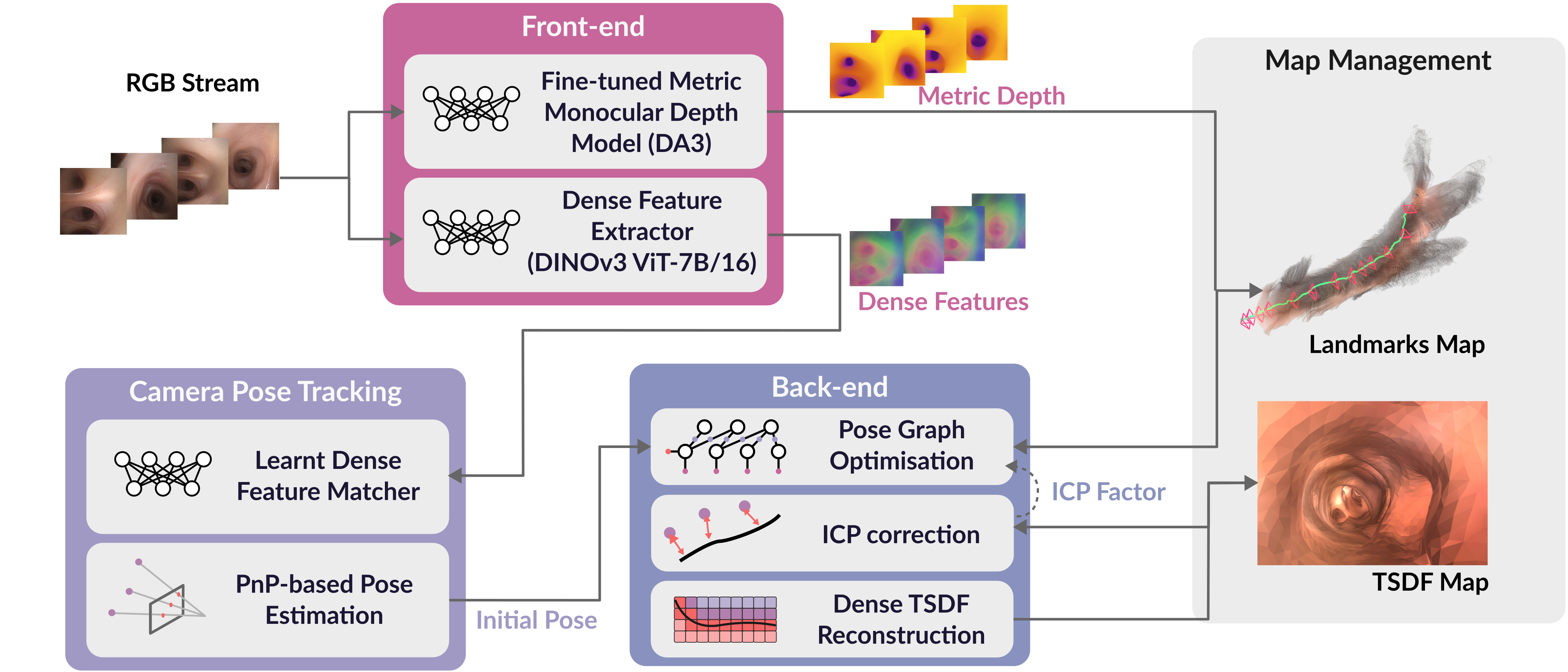}
  \caption{\textbf{System overview.} For every frame, the front end (Sec.~\ref{sec:matching}) predicts metric depth and dense descriptors. Camera pose tracking (Sec.~\ref{sec:pose}) matches descriptors across keyframes and solves PnP for an initial pose. The back end (Sec.~\ref{sec:backend}) refines that pose by pose-graph optimisation, registers the depth against the running map by ICP, adds the resulting constraint to the pose graph, and integrates the fused pose into a dense TSDF reconstruction. Map management (Sec.~\ref{sec:data_assoc}) maintains the resulting landmark and TSDF maps.}
  \label{fig:pipeline}
  \vspace{-10pt}
\end{figure*}

\figref{fig:pipeline} summarises our system. It takes monocular RGB bronchoscopy video with known intrinsics, and returns a metric trajectory of the scope tip together with a dense reconstruction of the endoluminal surface.

For every frame the front end predicts a metric depth map with per-pixel reliability from a bronchoscopy-adapted Depth Anything 3 (DA$^3$) backbone~\cite{lin2025depthanything3}, and extracts dense DINOv3 patch descriptors that stay discriminative on mucosa. The system keeps two maps: a sparse landmark map of 3D points back-projected from those measurements, and a dense surface map stored as a truncated signed distance field (TSDF). For each new keyframe, a trained matcher links the current descriptors to the landmarks predicted to be visible, and PnP on the surviving correspondences supplies an initial pose.
This pose is used to optimise the factor graph with the visual landmarks and motion model, and returns a visual-only pose. That pose initialises a projective point-to-plane ICP registration against the local TSDF submap, whose surviving correspondences re-enter the same graph for a second update. Only the fused pose are used for TSDF integration.

\subsection{Preliminaries}
\label{sec:notation}

Table~\ref{tab:notation} collects the symbols used throughout the paper.
The input is a sequence of monocular frames $\{\mathbf{I}_t\}_{t=0}^{T}$ with known intrinsics $\mathbf{K}$. We estimate the camera trajectory $\mathcal{X}=\{\pose{\W}{\C}^{t}\}_{t=0}^{T}$, a sparse landmark map $\mathcal{L}=\{\point{\ell}{\W}^{j}\}_{j=1}^{N}$, and a dense endoluminal surface stored as a TSDF $\Phi:\mathbb{R}^{3}\to\mathbb{R}$ on a voxel grid. The zero level set of $\Phi$ is the reconstructed airway wall.

\begin{table}[!t]
\centering
\captionsetup{font=footnotesize}
\caption{Notation used across the article.}
\label{tab:notation}
\footnotesize
\begin{tabular}{@{}ll@{}}
\toprule
\textbf{Symbol} & \textbf{Description} \\
\midrule
$\mathbf{I}_t$, $\mathbf{K}$ & RGB frame at time $t$; camera intrinsics \\
$\mathbf{D}_t$, $\mathbf{S}_t$ & predicted metric depth and its standard deviation \\
$\mathbf{Q}_t$ & per-pixel depth reliability, $\in[0,1]$ \\
$\pose{\W}{\C}^{t}$, $\pose{\W}{\C}^{t,\star}$ & camera pose; best camera pose estimate (after pass two) \\
$\point{\ell}{\W}^{j}$ & position of landmark $j$ in the world frame $\W$ \\
$\mathbf{u}_{tj}$, $z_{tj}$ & measured patch coordinate and optical-axis depth \\
$\sigma^z_{tj}$ & standard deviation of the depth measurement $z_{tj}$ \\
$\Phi$ & truncated signed distance field (TSDF) \\
\bottomrule
\end{tabular}
\end{table}

Poses follow the camera-to-world convention, $\pose{\W}{\C}^{t}\in\mathrm{SE}(3)$ and $\point{\ell}{\W}^{j}\in\mathbb{R}^{3}$, and landmark $j$ seen from camera $t$ has camera-frame coordinates $\left(\pose{\W}{\C}^{t}\right)^{-1}\point{\ell}{\W}^{j}=[x_{tj},y_{tj},z_{tj}]^T$. The calibrated projection model maps a camera-frame point to pixel coordinates,
\begin{equation}
    \pi_{\mathbf{K}}\big([x,y,z]^T\big) = \big[\,f_x x/z + c_x,\;\; f_y y/z + c_y\,\big]^T.
\label{eq:projection_model}
\end{equation}

\section{Front-end}
\label{sec:matching}

The front end produces three outputs per frame: metric depth with per-pixel reliability, a dense descriptor field, and the 3D landmark candidates that combine them.

\subsection{Monocular Metric Depth Estimation}
\label{sec:depth_frontend}
For every frame we predict a dense metric depth map $\mathbf{D}_t$ with the DA$^3$ backbone, fine-tuned on bronchoscopy renderings with ground truth depth. Fine-tuning adapts the predictor to the illumination and diameter of small airways and fixes its \emph{absolute} scale. This predictor is the only source of metric scale in the system.

A second head predicts a per-pixel depth uncertainty. We train it to output the scale $\mathbf{S}_t$, in millimetres, of a Laplace distribution over log-depth, because its heavier tails keep specular and occluded pixels, where depth error is largest, from dominating the fit. We combine $\mathbf{S}_t$ with two geometric terms and one temporal term into a per-pixel reliability
\begin{equation}
\mathbf{Q}_t = \exp\!\left[-\left(\frac{\mathbf{S}_t}{\tau_\sigma \mathbf{D}_t}\right)^{2}\right]\, q_r\, q_s\, q_\tau \;\in\; [0,1].
\label{eq:reliability}
\end{equation}
The leading factor divides the predicted error by the local depth, which makes it a \emph{fractional} error: one millimetre matters more at $5$\,mm range than at $50$\,mm. It approaches $1$ as that fraction vanishes and decays as it grows, and it is zero wherever the depth is invalid. $\mathbf{S}_t$ is a fixed fraction of $\mathbf{D}_t$ (\appref{app:depth_training}). The ratio therefore depends on the predicted uncertainty alone and does not grow with distance. Its decay constant is $\tau_\sigma=0.02$ on the synthetic phantom sequences, $0.05$ on the CT-derived airways and $0.15$ on ex vivo data, whose depth predictions are less accurate. $q_r$ discounts the tunnel centre, where monocular depth is least constrained. It is enabled on ex vivo data only. $q_s$ discounts patches whose descriptor resembles its immediate neighbours, since those are hard to re-find. $q_\tau$ discounts depths that disagree with the map already built, and applies once at least five mapped landmarks are in view.

$\mathbf{Q}_t$ is a unitless score. Sec.~\ref{sec:feature_frontend} converts it to a metric standard deviation for each pixel. The training data and optimisation schedule are given in \appref{app:depth_training}.

\subsection{Dense Features and Landmark Initialisation}
\label{sec:feature_frontend}
\label{sec:patch_measurements}
We extract dense image features from a frozen DINOv3 ViT-7B/16 backbone~\cite{simeoni2025dinov3}. It produces a $42 \times 42$ patch grid of $4096$-D descriptors from each $672 \times 672$ RGB frame. The matcher of Sec.~\ref{sec:image_to_map} projects these into a $256$-D matching space, and the map stores those vectors. We use DINOv3 because its features stay distinctive on textureless mucosa. Sec.~\ref{sec:ablation} substitutes ORB into the pipeline to measure the difference.

\begin{figure*}[t]
  \centering
  \includegraphics[width=0.95\linewidth]{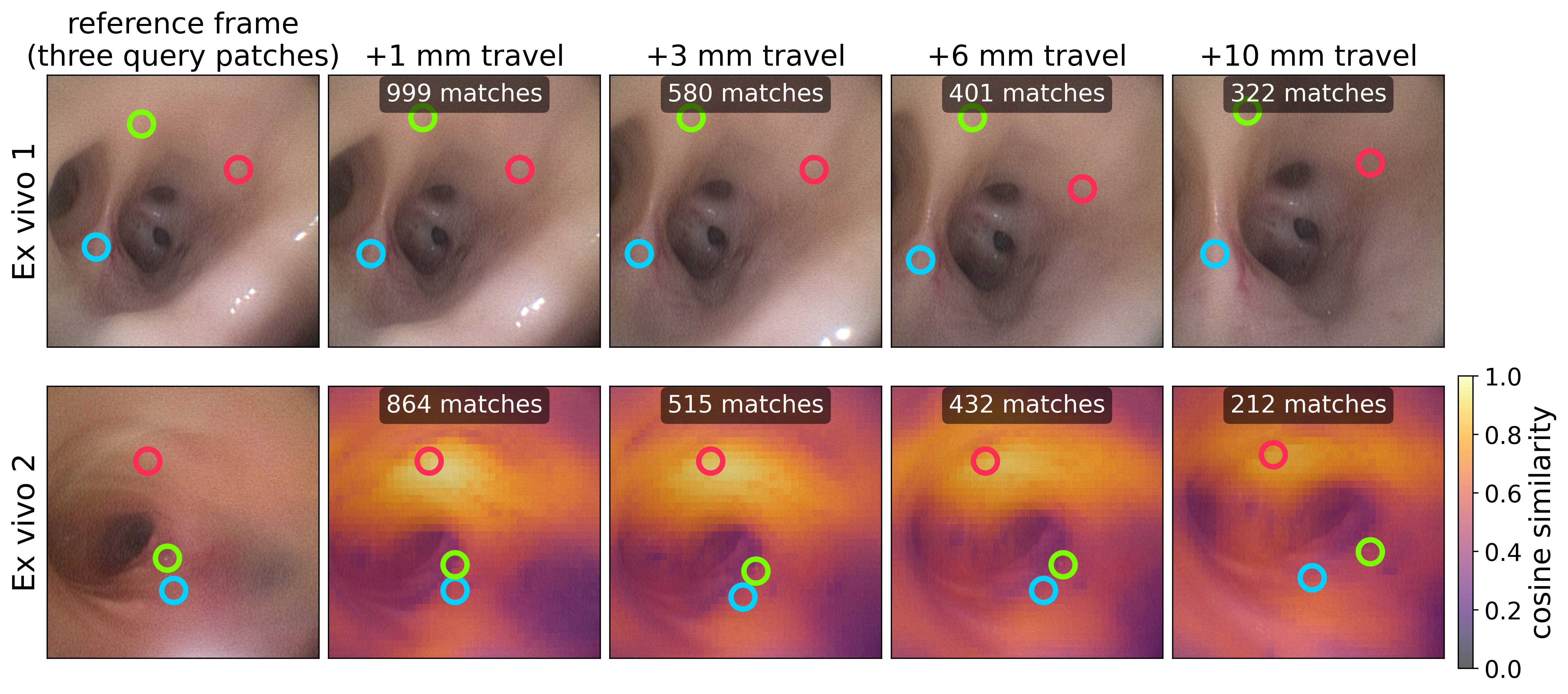}
  \caption{\textbf{Keypoint correspondences on textureless mucosa using DINO features.} The rows show two ex vivo sequences and the columns the same keypoints tracked over four forward displacements. The second row overlays the cosine similarity map for the red keypoint. The same regions are tracked despite the displacement. The front end therefore uses the pre-trained DINOv3 features without fine-tuning.}
  \label{fig:feature_similarity}
\end{figure*}

Fig.~\ref{fig:feature_similarity} shows an example of sparse keypoint tracking on two ex vivo sequences as the scope advances through the airway. A match is kept when the similarity of its best candidate exceeds that of the second-best candidate, or of the no-match option, by more than $0.05$ (the \emph{margin test}), and when tracking it back to the source frame lands within $16$\,px of its origin (the \emph{cycle test}). Several hundred of the $1764$ patches pass both tests at each separation, and $66$--$123$ per sequence pass at all four separations.

Depth and descriptors differ in resolution: a descriptor covers a $16\times16$ patch, whereas depth is predicted per pixel. We compute features on a resized, patch-aligned copy of the frame, so we scale the matcher's coordinates back to the original image before applying the intrinsics. A single $\mathbf{K}$ is then used everywhere and depth is not rescaled.

Each patch is assigned one depth and one reliability, both area-averaged over the pixels it covers. Patches with non-positive depth are discarded, as are those whose reliability $Q$ falls below $0.15$; that floor is held fixed across every sequence reported here. Back-projecting the centre of a surviving patch through $\mathbf{K}$ at its depth gives a 3D point, which inherits the patch descriptor.

The depth standard deviation converts the reliability of a patch into metric units,
\begin{equation}
\sigma^z_{tj} = \alpha\, z_{tj}\,\max\!\left(1-Q_{tj},\; \epsilon\right), \qquad \alpha = 0.1,\ \epsilon = 10^{-2},
\label{eq:depth_sigma}
\end{equation}
so it ranges from $0.1\,\%$ of the depth at $Q=1$ to $10\,\%$ at $Q=0$; the floor $\epsilon$ keeps the standard deviation from collapsing to zero. The depth factor constrains the camera-frame $z$ coordinate alone. Lateral position is left to the projection factor, whose covariance $\boldsymbol{\Sigma}^{\pi}_{tj}=(\sigma^{\pi})^{2}\mathbb{I}_{2}$ is isotropic.

\section{Map Creation and Management}
\label{sec:data_assoc}

We update the maps from the camera pose of each keyframe, which are estimated by the Camera Tracking process later described in Sec.~\ref{sec:pose}.

\subsection{Maps}
The sparse landmark map holds the 3D points that carry the visual factors. The dense map is the TSDF, which we integrate from posed depth frames. Because the integration uses the fused pose, Sec.~\ref{sec:tsdf_fusion} describes it with the back end.

Each landmark $j$ stores:
\begin{itemize}
  \item its position $\point{\ell}{\W}^{j}$, which is the variable the graph optimises;
  \item its five most recent descriptors, and the representative among them used for matching;
  \item the keyframes that have observed it;
  \item its state, \emph{tentative} or \emph{confirmed}.
\end{itemize}
A landmark starts as tentative and becomes confirmed at its second observation. Only confirmed landmarks are used for pose estimation (Sec.~\ref{sec:pose}).

\subsection{Map Initialisation}
\label{sec:map_init}
The first keyframe defines the world frame: we insert $\pose{\W}{\C}^{0}=\mathbb{I}_4$ and anchor it with a prior factor. Every reliable patch is back-projected,
\begin{equation}
    \point{\ell}{\W}^{j} = \pose{\W}{\C}^{0}\!\left(z_{0j}\,\mathbf{K}^{-1}\!\begin{bmatrix}\mathbf{u}_{0j}\\1\end{bmatrix}\right),
\label{eq:first_frame_backprojection}
\end{equation}
inserted as a landmark variable, and connected to $\pose{\W}{\C}^{0}$ by a projection factor and a depth factor; the projection factor alone would leave it under-constrained along its viewing ray. The factors are defined in Sec.~\ref{sec:backend}.

\subsection{Updating the Map with New Observations}
For a new keyframe $t$, the predicted pose of Sec.~\ref{sec:pose} selects candidate landmarks by field of view, positive depth, and TSDF occlusion. Our matcher (Sec.~\ref{sec:image_to_map}) then proposes correspondences between those candidates and the current patches. We filter proposals by descriptor tests, visibility and depth consistency, AP3P RANSAC, and a projection-residual check after optimisation. Only accepted matches create observation records and descriptor updates.

\subsection{Landmark Birth}
Patches that match no existing landmark are considered for birth once the current pose has been accepted and optimised. Birth requires the reliability and depth tests of Sec.~\ref{sec:patch_measurements}, a keyframe labelled stable (Sec.~\ref{sec:pose}), and a duplicate check against nearby landmarks. We initialise the new landmark with~\eqref{eq:first_frame_backprojection} evaluated at the accepted pose, and it enters the graph as in Sec.~\ref{sec:map_init}.

The map stores the $256$-D descriptors that the matcher compares (the projection of the $4096$-D DINOv3 vectors), and matches against whichever of a landmark's five lies closest to their mean. When a landmark is re-observed, we average the new descriptor with that representative at rate $0.1$, re-normalise it, and add it to the bank in place of the raw one. This lets the stored appearance follow gradual change.

\section{Camera Pose Tracking}
\label{sec:pose}

We localise each new keyframe against the map in two steps: descriptor matching gives 2D--3D correspondences, and PnP gives an initial pose. The factor graph of Sec.~\ref{sec:backend} then refines it together with the map.

\subsection{Image-to-Map Matching}
\label{sec:image_to_map}
The landmarks predicted to be visible (Sec.~\ref{sec:data_assoc}) are matched with the current frame's patches at sub-pixel accuracy, using a custom-trained cross-attention matcher. Its architecture, training data, and losses are described in \appref{app:matcher_training}.

Matches are accepted by the margin and cycle tests of Sec.~\ref{sec:feature_frontend} and carry the projection-noise model given there. On the ex vivo sequences we choose the reference keyframes by estimated pixel displacement, targeting separations of $0$, $15$ and $30$\,px. Close references give reliable matches, and distant ones support triangulation. 

\subsection{Pose Initialisation}
\label{sec:pose_init}
\noindent\textbf{PnP.} Given the landmark correspondences and the current graph estimates $\point{\ell}{\W}^{j}$, AP3P RANSAC ($3000$ iterations, $3.0$\,px reprojection threshold) estimates a pose hypothesis, refined on its inliers. An inlier floor of $\min(25,\max(20,\,0.08\,n_{\text{raw}}))$ rejects solutions backed by too few correspondences. The floor rises from $20$ to $25$ as the candidate count grows. If RANSAC fails, iterative PnP on the same correspondences is the fallback, with the landmark coordinates held at their current graph estimates. If both fail, a constant-velocity model predicts the pose, and a weak between-pose factor ties the new keyframe to the previous one.

\noindent\textbf{Pose quality.} A keyframe is marked \emph{stable} only when the primary RANSAC pose estimate succeeds with enough inliers. Only stable keyframes extend the map: the others keep the trajectory connected, but cannot create landmarks, update descriptors, or drive TSDF integration. We loosen the between-pose factor as the pose it follows gets weaker, from $5$\,mm and $2^\circ$ after a RANSAC pose to $50$\,mm and $10^\circ$ after a constant-velocity one. On the ex vivo sequences, backward motion beyond $2$\,mm along the optical axis is rejected and replaced by the velocity prediction.

\section{Back-end}
\label{sec:backend}

\subsection{Factor Graph Formulation}
The back end estimates the camera trajectory $\mathcal{X}$ and the landmark map $\mathcal{L}$ jointly, treating every front-end output as a measurement in a single factor graph. It is solved in two passes: the first uses the visual landmarks alone, the second adds a geometric factor from point-to-plane ICP against the surface already reconstructed, constraining the motion that is perpendicular to the airways' direction.

\vspace{0.5em}
\noindent\textbf{Reprojection factor.} Every accepted observation ties landmark $j$ to keyframe $t$ through the residual
\begin{equation}
    \mathbf{r}_{\text{proj}} = \pi_{\mathbf{K}}\big( \left(\pose{\W}{\C}^t \right)^{-1} \point{\ell}{\W}^{j} \big) - \mathbf{u}^{t, j},
\label{eq:projection_residual}
\end{equation}
where $\pi_{\mathbf{K}}$ is the camera projection function, and $\mathbf{u}^{t,j}$ is the measured patch coordinate, under the isotropic image covariance $\boldsymbol{\Sigma}^{t,j}_{\text{proj}}$ of Sec.~\ref{sec:patch_measurements}.

\vspace{0.5em}
\noindent\textbf{Depth factor.} For each landmark we also define a depth residual that constrains the depth to the estimated value (from the fine-tuned DA3 model):
\begin{equation}
    \mathbf{r}_{\text{depth}} = \big[\left( \pose{\W}{\C}^t \right)^{-1} \point{\ell}{\W}^{j} \big]_z - z^{t,j},
\label{eq:depth_residual}
\end{equation}
where $z^{t,j}$ is the corresponding depth estimate of the pixel $j$ from keyframe $t$ $\mathbf{D}_t$, with standard deviation $\sigma^{t,j}_{z}$ of~\eqref{eq:depth_sigma}. This residual constrains the $z$ coordinate in the camera frame (the optical-axis depth), which differs from the Euclidean range. Later observations of the same landmark carry the reprojection residual alone.

\vspace{0.5em}
\noindent\textbf{Anchor factor.} A unary prior on the first pose fixes the global position and orientation that a monocular graph cannot otherwise determine. Writing $\bar{\mathbf{X}}_0$ for the anchor, held at the identity at frame zero, its residual is
\begin{equation}
    \mathbf{r}_{\text{anchor}} = \mathrm{Log}\big( \pose{\W}{\C}^{0}\big),
\label{eq:gauge_residual}
\end{equation}
with a fixed covariance $\boldsymbol{\Sigma}^0$. 

\vspace{0.5em}
\noindent\textbf{Motion factor.} A weak between-pose factor encodes the bronchoscope's forward advance. With $\Delta \nmat{T}^{t-1,t}\in\mathrm{SE}(3)$ the relative motion predicted by the constant-velocity model of Sec.~\ref{sec:pose}, it contributes
\begin{equation}
    \mathbf{r}_{\text{motion}} = \mathrm{Log}\left(\left(\Delta \pose{\W}{\C}^{t-1}\, \nmat{T}^{t-1,t}\right)^{-1} \pose{\W}{\C}^{t}\right),
\label{eq:motion_residual}
\end{equation}
under a covariance $\boldsymbol{\Sigma}^{m,t}$ set by the predecessor's pose-quality label (Sec.~\ref{sec:pose}).

\begin{figure}[!t]
  \centering
  \captionsetup{font=footnotesize}
  \includegraphics[width=0.95\linewidth]{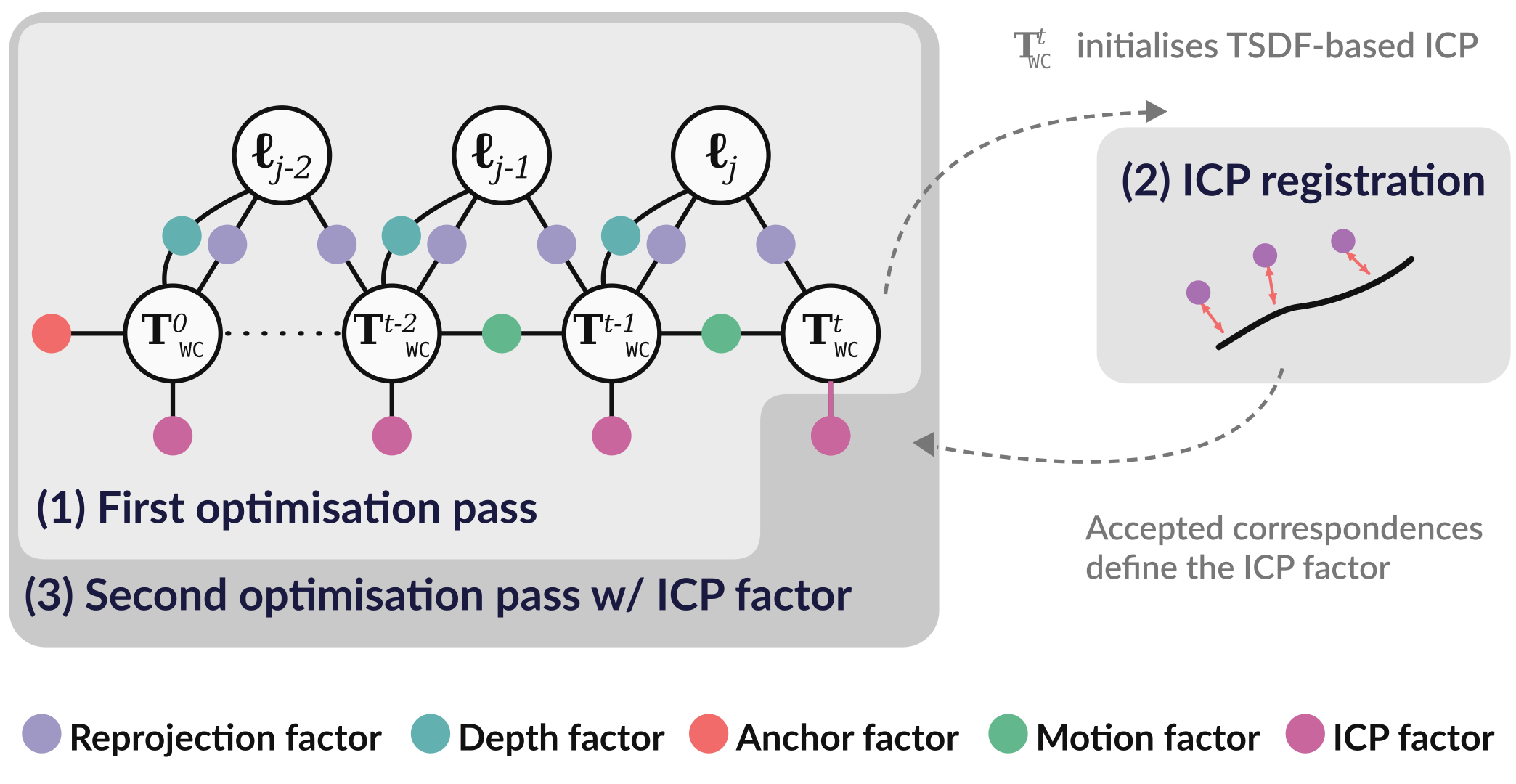}
  \caption{\textbf{Two-pass factor graph.} White circles are keyframe poses $\pose{\W}{\C}^{t}$ and landmarks $\point{\ell}{\W}^{j}$; coloured circles are factors, one colour per type (legend, bottom). \textbf{(1)}~The first pass optimises an anchor factor on the first pose, a motion factor between consecutive poses, a reprojection factor for each observation of a landmark, and one depth factor between a landmark and the keyframe that created it. It returns the first-pass pose $\pose{\W}{\C}^{t}$. \textbf{(2)}~That pose initialises the point-to-plane ICP registration of Sec.~\ref{sec:icp} against the TSDF submap. \textbf{(3)}~The accepted correspondences add one ICP factor on the current pose, alongside the ICP factors of earlier poses, and re-optimising propagates the correction to earlier poses and landmarks, returning the fused pose $\pose{\W}{\C}^{t,\star}$.}
  \label{fig:factor_graph}
\end{figure}

\vspace{0.5em}
\noindent\textbf{ICP factor.} At every timestep $t$ we optimise the graph twice. The first pass uses the factors above and the history, and returns the pose $\pose{\W}{\C}^{t}$. We then register the current depth against the TSDF map by point-to-plane ICP, initialised from this pose (Sec.~\ref{sec:icp}), and the second pass adds one ICP factor on $\pose{\W}{\C}^{t}$ (\figref{fig:factor_graph}). The factor stacks one residual for each correspondence $i$ accepted by ICP,
\begin{equation}
    r_{\text{icp},i} = \mathbf{n}_i^{\top}\big( \pose{\W}{\C}^{t}\,\mathbf{p}_i - \mathbf{q}_i \big),
\label{eq:icp-factor}
\end{equation}
where $\mathbf{p}_i$ is the point in the camera frame, and $\mathbf{q}_i$ and $\mathbf{n}_i$ are the corresponding TSDF surface point and normal in the world frame. We weight each residual by $\sigma^{\mathrm{icp}}_i$ of~\eqref{eq:icp_sigma}.

\vspace{0.5em}
\noindent\textbf{Factor graph optimisation.} We solve the estimation problem using a factor graph formulation. The second pass minimises the objective below, and the first pass omits its last term:
\begin{equation}
\begin{aligned}
    \min_{\mathcal{X},\mathcal{L}}\quad
    & \| \mathbf{r}_{\text{anchor}} \|^2_{\boldsymbol{\Sigma}^0}
    + \sum_{(t,j)\in\mathcal{O}} \rho \left( \| \mathbf{r}_{\text{proj}} \|^2_{\boldsymbol{\Sigma}_{\text{proj}}^{t,j}} \right)\\
    &+ \sum_{(t,j)\in\mathcal{O}_z} \rho \left(( \| \mathbf{r}_{\text{depth}} \| ^{2}_{ \sigma_z^{t,j}} \right)
    + \sum_{t\in\mathcal{M}} \rho \left( \| \mathbf{r}_{\text{motion}} \|^2_{\boldsymbol{\Sigma}^{m,t}} \right)\\
    & + \sum_{t\in\mathcal{D}} \sum_{i\in\mathcal{C}_t} \rho \left( r_{\text{icp},i}^{2} / (\sigma_{\mathrm{icp}}^{i})^{2} \right),
\end{aligned}
\label{eq:full_graph_objective}
\end{equation}
where $\rho_{(\cdot)}$ is a robust kernel for each factor, 
$\mathcal{O}$ is the set of landmark observations and $\mathcal{O}_z\subseteq\mathcal{O}$ those that also carry a depth residual. $\mathcal{M}$ is the set of frames carrying a motion factor, and $\mathcal{D}$ those carrying an ICP factor, whose accepted correspondences are $\mathcal{C}_t$. We solve this problem using iSAM2 ~\cite{kaess2012isam2} for each incoming keyframe.

\subsection{Point-to-Plane ICP}
\label{sec:icp}
The first-pass pose rests on a few hundred patch correspondences, which can under-constrain the camera motion. We therefore register the dense depth prediction against a local TSDF map of the last $15$ keyframes and let the surviving correspondences enter the graph as an extra constraint, improving both trajectory and surface accuracy (Sec.~\ref{sec:ablation}). The registration is standard projective point-to-plane ICP, as in KinectFusion~\cite{newcombe2011kinectfusion}: ray-cast correspondences, a robust M-estimator and a damped Gauss-Newton solve, run here against the TSDF with a custom GPU implementation.

\vspace{0.5em}
\noindent\textbf{Data association.} We raycast each pixel of the current view to the surface. Along the ray we sample the trilinearly interpolated field every $\delta=0.8\,s_v$, for voxel size $s_v$. At the first sign change, from $\Phi_k>0$ to $\Phi_{k+1}\leq 0$, we place the surface point the fraction $\Phi_k/(\Phi_k-\Phi_{k+1})$ of a step past sample $k$. 
Normals come from the TSDF gradient by central differences with trilinear interpolation. 

\vspace{0.5em}
\noindent\textbf{Robust point-to-plane ICP.} Let $\mathbf{p}_i$ be a source point in the current camera frame and $(\mathbf{q}_i,\mathbf{n}_i)$ its corresponding TSDF surface point and normal in the submap, with accumulated weight $W(\mathbf{q}_i)$ of~\eqref{eq:tsdf_update_w}. With $\pose{\A}{\C}^{t}=\big(\pose{\W}{\A}\big)^{-1}\pose{\W}{\C}^{t}$ the transform from the current camera to the submap, the point-to-plane residual is
\begin{equation}
r_i\big(\pose{\A}{\C}^{t}\big) = \mathbf{n}_i^T\!\left(\pose{\A}{\C}^{t}\,\mathbf{p}_i - \mathbf{q}_i\right),
\label{eq:icp_residual}
\end{equation}
with the standard rotation-first Jacobian $\mathbf{J}_i$ under a left SE(3) perturbation.

Specular highlights and depth artefacts produce outliers, so we minimise the residuals under a Welsch kernel with graduated non-convexity~\cite{yang2020graduated}, modulating its scale $\tau$ from wide to tight,
\begin{equation}
\rho_\tau(r) = \tau^2\!\left(1 - \exp\!\left(-\tfrac{r^2}{2\tau^2}\right)\right), \qquad \tau_k = \tau_0 \, \gamma^k,
\label{eq:welsch}
\end{equation}
with $\tau_0 = \max(P_{95}(|r_i|), 3\sigma_{\text{med}})$, $\gamma = (\tau_{\min}/\tau_0)^{1/n_\gamma}$, $n_\gamma=4$ stages, $\tau_{\min} = \max(\tau_{\text{floor}},\sigma_{\text{med}})$ with floor $\tau_{\text{floor}}=1\,\mu\text{m}$, and $\sigma_{\text{med}}$ the median of $\sigma^{\mathrm{icp}}_i$ (below) over the current correspondences. The IRLS weights are $w_i = \beta_i\exp(-r_i^2/(2\tau^2))$, where
\begin{equation}
\beta_i = \frac{\cos\theta_i}{(\sigma^{\mathrm{icp}}_i)^2}, \qquad
\sigma^{\mathrm{icp}}_i = \sqrt{\mathbf{S}_t(\mathbf{u}_i)^2 + \dfrac{s_v^2}{\max(W(\mathbf{q}_i),1)}}
\label{eq:icp_sigma}
\end{equation}
combines the DA$^3$ depth precision $\mathbf{S}_t$ with the TSDF's own accumulated evidence at that voxel and the cosine of the surface incidence angle. Each stage solves the damped normal equations
\begin{equation}
\!\big(\mathbf{J}^T \boldsymbol{\Omega}_k \mathbf{J} + \lambda_{\mathrm{LM}}\,\mathrm{diag}(\mathbf{J}^T \boldsymbol{\Omega}_k \mathbf{J})\big)\delta\boldsymbol{\xi} = -\mathbf{J}^T \boldsymbol{\Omega}_k \mathbf{r},
\label{eq:normal_equations}
\end{equation}
with $\boldsymbol{\Omega}_k = \mathrm{diag}(w_1,\ldots,w_N)$ the stage-$k$ weights, $\mathbf{r}$ and $\mathbf{J}$ stacked over correspondences, and $\lambda_{\mathrm{LM}}$ the Levenberg--Marquardt damping. A backtracking line search ($\eta \in \{1,0.5,0.25\}$) controls step acceptance; damping resets each stage and the schedule advances after three consecutive rejections.

\vspace{0.5em}
\noindent\textbf{Anisotropic conditioning.} Write $\mathbf{H} = \mathbf{J}^T \boldsymbol{\Omega}_k \mathbf{J} + \lambda_{\mathrm{LM}}\,\mathrm{diag}(\mathbf{J}^T \boldsymbol{\Omega}_k \mathbf{J})$ for the matrix on the left of~\eqref{eq:normal_equations}. It couples rotation (radians) and translation (millimetres) at incomparable scales, so we apply a diagonal scaling $\mathbf{G} = \mathrm{diag}(1,1,1,d_{\text{med}},d_{\text{med}},d_{\text{med}})$, where $d_{\text{med}}$ is the median source depth, and take the eigendecomposition of the scaled matrix, $\mathbf{H}_s = \mathbf{G}^T \mathbf{H} \mathbf{G} = \mathbf{V}\boldsymbol{\Lambda}\mathbf{V}^T$, with orthonormal eigenvectors $\mathbf{V}$ and eigenvalue diagonal $\boldsymbol{\Lambda}=\mathrm{diag}(\lambda_i)$. The spectrum reports how strongly the surface constrains each direction: inside a tube the eigenvalue along the lumen axis is orders of magnitude smaller than the radial ones. We use it to diagnose degeneracy and to compute the update without altering it, giving the update
\begin{equation}
\delta\boldsymbol{\xi} = \mathbf{G}\,\mathbf{V}\,\boldsymbol{\Lambda}^{-1}\mathbf{V}^T(-\mathbf{G}^T \mathbf{b}), \qquad \mathbf{b}=\mathbf{J}^T\boldsymbol{\Omega}_k\mathbf{r},
\label{eq:conditioned_update}
\end{equation}
with $\mathbf{b}$ the gradient. When the spectrum stays ill-conditioned, we discard the ICP correction for that keyframe and keep the pose the visual factors already provide.

\vspace{0.5em}
\noindent\textbf{Solving and accepting the alignment.} We initialise the alignment using the estimate of $\pose{\W}{\C}^{t}$ obtained from the first optimisation pass (\figref{fig:factor_graph}). The solver minimises
\begin{equation}
\bar{\nmat{T}}_{\A\C}^{t} = \arg\min_{\mathbf{T}} \sum_i \beta_i\, \rho_\tau\!\big(r_i(\mathbf{T})\big),
\label{eq:icp_objective}
\end{equation}
with the reliability weights $\beta_i$ above, and returns the refined pose $\bar{\nmat{T}}_{\W\C}^{t,\mathrm{icp}} = \pose{\W}{\A}\,\bar{\nmat{T}}_{\A\C}^{t}$ in the world frame. 
We add an ICP factor only if the ICP result passes three conditions.
\begin{enumerate}[label=(\roman*),nosep]
\item \emph{Support.} The registration returns at least $100$ correspondences.
\item \emph{Convergence.} The solver converges.
\item \emph{Rotation consistency.} A converged registration with at least $200$ correspondences must not differ from the first-pass rotation by more than $15^\circ$. Only rotation about directions that the ICP Hessian constrains counts, because the roll about a straight lumen is unobservable. Beyond the threshold the first-pass pose was confidently wrong, and we reject the keyframe.
\end{enumerate}
A keyframe that fails (i) or (ii) loses only its surface constraint, and tracking continues on the visual factors. An accepted registration passes the graph its correspondences at the converged pose, expressed in the world frame, $(\mathbf{p}_i,\mathbf{q}_i,\mathbf{n}_i,\sigma^{\mathrm{icp}}_i)$, as the residuals of~\eqref{eq:icp-factor}.

\subsection{TSDF Fusion}
\label{sec:tsdf_fusion}

The surface is built from the output of the factor graph optimisation $\pose{\W}{\C}^{t,\star}$, the most refined estimate the system produces. Integration happens once per keyframe, after the second pass, and only when the keyframe is labelled \emph{stable} (Sec.~\ref{sec:pose}).

For a voxel centre $\mathbf{x}$, define $\mathbf{p}_c=\big(\pose{\W}{\C}^{t,\star}\big)^{-1}\mathbf{x}$ and $\mathbf{u}=\pi_{\mathbf{K}}(\mathbf{p}_c)$. When $\mathbf{u}$ lies inside the image and $D_t(\mathbf{u})$ is valid, let $\mathbf{p}^{s}_t$ be the back-projection of $\mathbf{u}$ at depth $D_t(\mathbf{u})$. The signed distance is the difference between the range to the observed surface and the range to the voxel, along that viewing ray,
\begin{align}
d_t(\mathbf{x}) &= \|\mathbf{p}^{s}_t\| - \|\mathbf{p}_c\|,
\label{eq:tsdf_signed_distance}\\
\phi_t(\mathbf{x}) &= \mathrm{clip}\!\left(d_t(\mathbf{x})/\mu,\,-1,\,1\right),
\label{eq:tsdf_observation}
\end{align}
where $\mu$ is the truncation distance; voxels with $|d_t(\mathbf{x})|>\mu$ are not updated. The running TSDF average obeys
\begin{align}
\Phi^{+}(\mathbf{x}) &= \frac{W(\mathbf{x})\Phi(\mathbf{x}) + w_t(\mathbf{x})\phi_t(\mathbf{x})}{W(\mathbf{x})+w_t(\mathbf{x})},
\label{eq:tsdf_update_phi}\\
W^{+}(\mathbf{x}) &= W(\mathbf{x})+w_t(\mathbf{x}).
\label{eq:tsdf_update_w}
\end{align}
Here $W(\mathbf{x})$ is the accumulated evidence weight. Integration into a voxel stops once $W(\mathbf{x})$ reaches a cap $W_{\max}$, which freezes well-observed geometry against later, noisier evidence. We weight each pixel by the precision the depth predictor reports for it,
\begin{equation}
w_t(\mathbf{x}) = \mathrm{clip}\!\left(\frac{1}{S_t(\mathbf{u})^{2}},\; w_{\min},\, w_{\max}\right),
\label{eq:tsdf_weight}
\end{equation}
using the standard deviation $\mathbf{S}_t$ of Sec.~\ref{sec:depth_frontend}, because a millimetre of depth error means the same thing at every voxel, whereas the reliability $\mathbf{Q}_t$ governs only the sparse factors. The clip stops a single confident but wrong pixel from dominating a voxel. Pixels with invalid depth, depth beyond the integration range, or a failed validity mask are not integrated.

A per-voxel consistency check protects established geometry. Once a voxel has accumulated weight above $3$, we attenuate by $0.05\times$ any incoming observation whose discrepancy $|d_t(\mathbf{x})-\mu\,\Phi(\mathbf{x})|$ exceeds $2$\,mm, and repeated agreeing evidence still corrects the voxel gradually. The ex vivo configuration uses $5$, $4$\,mm and $0.1\times$.

\section{Implementation Details}
\label{sec:implementation}

The system is implemented in Python. The front end builds on PyTorch modules, and the back end uses the Python bindings of the GTSAM library. The dense TSDF and the ICP solver are custom CUDA kernels. The optimisation problem is implemented with GTSAM and runs on the CPU.

The TSDF uses $0.5$\,mm voxels on the synthetic sequences and $1.0$\,mm on ex vivo data, with a truncation distance of four voxels. We extract meshes with marching cubes on its zero level set. Projection factors use isotropic image noise, $\sigma=2$\,px synthetic and $4$\,px ex vivo, under a Huber kernel. Depth factors use the standard deviation of~\eqref{eq:depth_sigma}, and each ICP residual uses $\sigma^{\mathrm{icp}}_i$ of~\eqref{eq:icp_sigma} under a Huber kernel.

\FloatBarrier

\section{Experiments}
\label{sec:experiments}

\newcommand{\hm}[2]{\cellcolor{hm#1}{#2}}

\subsection{Setup}
\label{sec:setup}

\begin{figure}[!t]
  \centering
  \includegraphics[width=0.95\columnwidth]{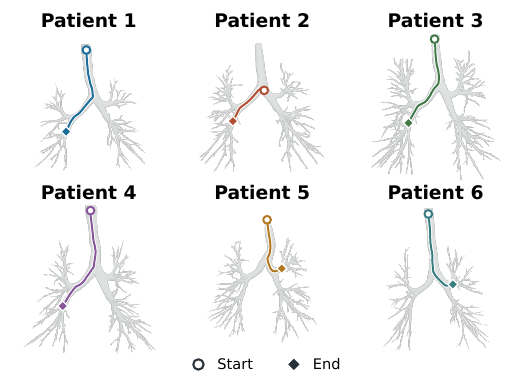}
  \caption{\textbf{Camera paths on the six CT-derived airways.} Each panel is a synthetic airway rendered from the CT scan of one patient (all six are ATM'22 cases~\cite{zhang2023atm22}; Sec.~\ref{sec:setup}). Open circles mark the start; filled diamonds mark the end.}
  \label{fig:branch_preview}
\end{figure}

\noindent\textbf{Datasets.} We use three sets and refer to them by these names throughout.
\begin{enumerate}[label=(\roman*),nosep,leftmargin=*,align=left, widest=iii]
\item \emph{CT-derived airways} (Patients 1 to 6). Six synthetic bronchoscopy sequences that we render from patient lung CT scans. Patients 1 to 6 are cases ATM\_001, ATM\_003, ATM\_004, ATM\_007, ATM\_124 and ATM\_063 of the public multi-site ATM'22 dataset~\cite{zhang2023atm22}. The airways span the airway tree from primary bronchi ($8$--$15$\,mm diameter) down to subsegmental branches ($2$--$4$\,mm), as shown in Fig.~\ref{fig:branch_preview}. They cover the difficulties of Sec.~\ref{sec:introduction}: long cylindrical segments, textureless walls, and narrow bifurcations with sharp viewpoint change and self-occlusion. We render RGB sequences with endoscopic lighting and procedural texture variation, each traversing a single branch. The six comprise $3{,}376$ frames, between $312$ and $714$ per sequence, and $90$--$203$\,mm of catheter travel. Ground truth depth and pose come from the simulator and are used only for evaluation. On these sequences the predicted depth and its uncertainty are multiplied by one factor, $0.857$, the geometric mean of the ratio of ground truth to predicted depth over ten further CT-derived airways that are not among the six. This set is the benchmark of Exp.~1 and Exp.~2.
\item \emph{Synthetic phantom sequences} B00, B46, B71, B96, B97 and B118, named by branch index. They are six branches of the same phantom lung and are used only for the ablation of Exp.~3.
\item \emph{Ex vivo sequences.} Sequences from ex vivo lung procedures with a custom continuum robot carrying an integrated RGB camera and co-located light source (Sec.~\ref{sec:real_bronch}). Two of them have no ground truth pose. The others carry an electromagnetic tracking (EMT) sensor, which provides a ground truth position.
\end{enumerate}
The synthetic sequences are static; the ex vivo lungs are ventilated and breathe during acquisition. Sec.~\ref{sec:limitations} discusses the rigid-scene assumption this leaves in place.

\vspace{0.5em}
\noindent\textbf{Baselines.} We compare our system against five recent monocular SLAM systems that use different tracking strategies: classical sparse feature-based (ORB-SLAM3~\cite{campos2021orbslam3}); a surgical-domain feed-forward reconstructor trained on endoscopic video (Endo3R~\cite{guo2025endo3r}); and three general-domain feed-forward systems (MASt3R-SLAM~\cite{murai2025mast3rslam}, VGGT-SLAM 2.0~\cite{maggio2026vggtslam2}, ViSTA-SLAM~\cite{zhang2025vistaslam}). Endo3R is the only baseline whose training distribution already contains endoluminal tissue. All methods receive the same rendered sequences; the baselines consume every frame, while our pipeline processes every third frame on these sequences. Baselines use the best parameter configuration found by grid search on the training sequences.

\noindent\textbf{Evaluation.} To evalaute the performance of our method, we report trajectory ATE and three reconstruction metrics on runs with more than $80\,\%$ temporal coverage; runs below that threshold are considered as failed.

The \emph{Absolute Trajectory Error} (ATE, RMSE in mm) measures the average trajectory error of the estimated camera trajectories after a Sim(3) alignment (SE(3) alignment for our method) over the first 40\,\% of matched poses. The ground truth of the synthetic sequence is obtained from the simulator; the ground truth of part of the ex-vivo sequences is provided by the EMT.

To evaluate the reconstructions, we produce a ground truth cloud for the synthetic sequences by back projecting the ground truth depth at the ground truth trajectory poses, after we remove invalid far-plane samples. The estimated clouds produced by each method are transformed by the same alignment of the trajectories.  We crop the estimated clouds to fit the ground truth evaluation volume and voxelise them at $1$\,mm to compute the error metrics we present as follows.

The \emph{F-score @ 10\,mm} (\%)~\cite{knapitsch2017tanks} is the harmonic mean of precision and recall at a $10$\,mm tolerance. \emph{Chamfer distance} (CD) is the RMS symmetric nearest-neighbour distance and measures typical surface error. The \emph{Hausdorff distance} (HD)~\cite{huttenlocher1993hausdorff} is the symmetric worst-case nearest-neighbour distance, $\mathrm{HD}=\max\{\sup_{p\in P}\inf_{g\in G}\|p-g\|,\;\sup_{g\in G}\inf_{p\in P}\|g-p\|\}$, the largest distance by which the two surfaces disagree~\cite{taha2015metrics}.

\subsection{Exp 1 -- Metric Depth and Scale Accuracy}
\label{sec:depth_accuracy}

\begin{table}[t]
\centering
\footnotesize
\renewcommand{\arraystretch}{1.0}
\setlength{\tabcolsep}{6pt}
\caption{\textbf{Metric depth accuracy} of the bronchoscopy-adapted DA$^3$ predictor on the six CT-derived airways (Patients 1 to 6, Sec.~\ref{sec:setup}), per valid pixel against simulator ground truth. The scale factor $s^\star=\mathrm{median}(z/\hat z)$ tests absolute scale. Lower AbsRel/RMSE, higher $\delta_1$, and $s^\star$ nearer $1.00$ are better.}
\label{tab:depth}
\begin{tabular}{l c c c c}
\toprule
\textbf{Anatomy} & \textbf{AbsRel} $\downarrow$ & \textbf{RMSE (mm)} $\downarrow$ & $\boldsymbol{\delta_1}$ \textbf{(\%)} $\uparrow$ & $\boldsymbol{s^\star}$ ($\to 1$) \\
\midrule
Patient 1 & $0.174$ & $5.22$ & $73.5$ & $0.883$ \\
Patient 2 & $0.186$ & $4.80$ & $75.3$ & $0.882$ \\
Patient 3 & $0.159$ & $5.13$ & $79.3$ & $0.954$ \\
Patient 4 & $0.116$ & $3.65$ & $88.3$ & $1.058$ \\
Patient 5 & $0.171$ & $4.98$ & $75.4$ & $0.931$ \\
Patient 6 & $0.148$ & $4.50$ & $85.7$ & $0.920$ \\
\midrule
\textbf{Mean} & $\mathbf{0.159}$ & $\mathbf{4.71}$ & $\mathbf{79.6}$ & $\mathbf{0.938}$ \\
\bottomrule
\end{tabular}
\end{table}

We first evaluate the scale accuracy of the depth predictor, since it provides scale for the whole reconstruction pipeline. The depth factor of~\eqref{eq:depth_residual} and the TSDF integration of Sec.~\ref{sec:tsdf_fusion} both use the predicted depth output, so a systematic scale error would propagate to the trajectory and the surface alike.

We evaluate the predictor directly on the six synthetic airways, where the simulator supplies ground truth depth (Sec.~\ref{sec:setup}). Table~\ref{tab:depth} reports, per valid pixel, the absolute relative error (AbsRel) and RMSE (mm), the threshold accuracy $\delta_1$ ($\max(\hat z/z,\,z/\hat z) < 1.25$), and the global scale factor $s^\star=\mathrm{median}(z/\hat z)$. $s^\star$ is computed on the depth the pipeline consumes, after the factor of Sec.~\ref{sec:setup} and without per-frame rescaling, to test the metric-depth assumption rather than relative depth quality.

Across the six airways, the predictor reaches a mean AbsRel of $0.159$ and $\delta_1$ of $79.6\,\%$, with an RMSE of $4.71$\,mm over a working range of a few tens of millimetres. The quantity that governs the reconstruction is $s^\star$: its mean is $0.938$ and the six span $0.882$ to $1.058$. The evaluation uses every valid pixel of every frame of each traverse, masking non-positive ground truth and pixels beyond the predictor's trained range.

\begin{figure*}[t]
  \centering
  \includegraphics[width=0.92\linewidth]{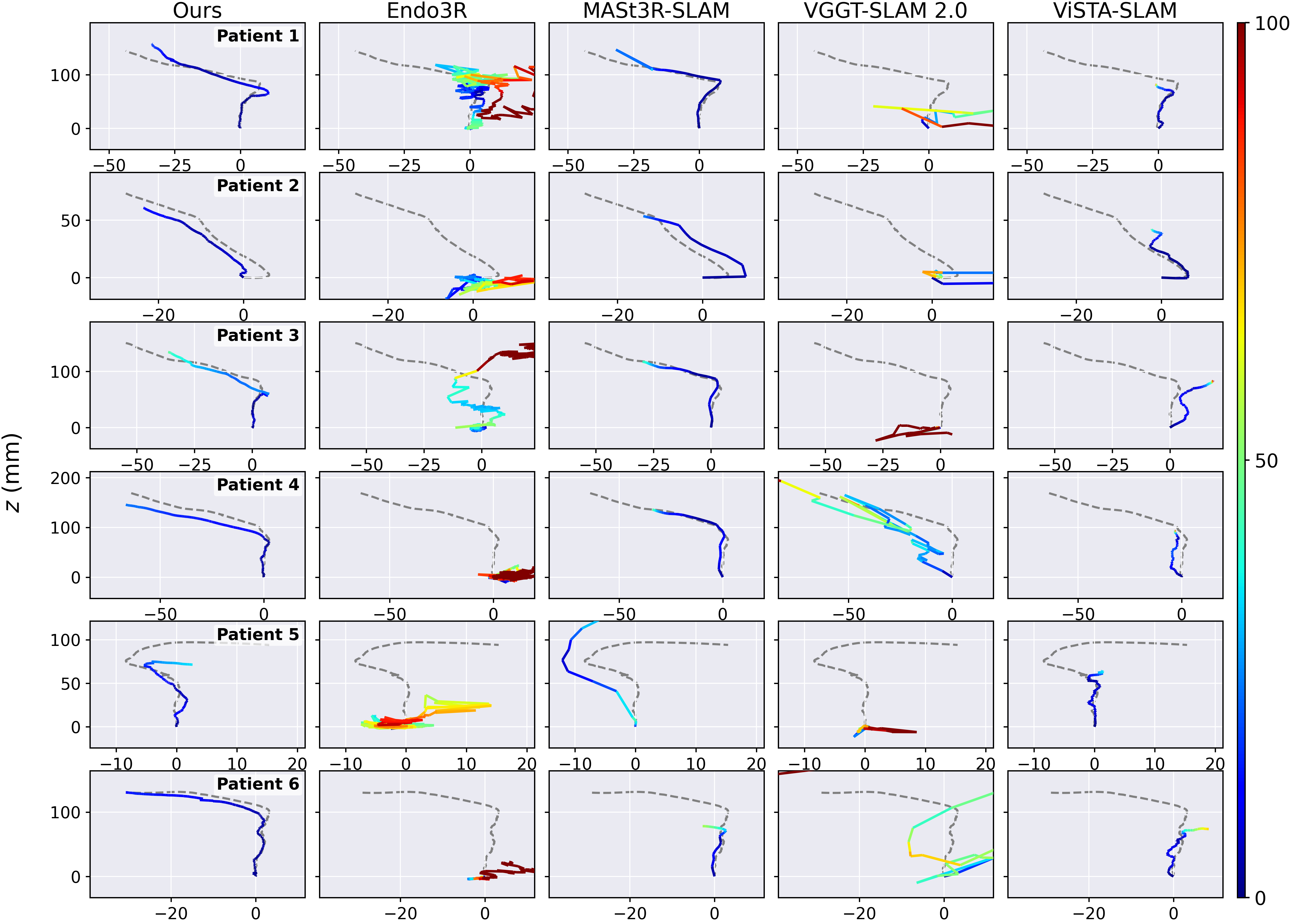}
  \caption{\textbf{Estimated trajectories on the six CT-derived airways} (rows: synthetic airways rendered from the CT scans of Patients 1 to 6, all ATM'22 cases~\cite{zhang2023atm22}; Sec.~\ref{sec:setup}) for ours, Endo3R, MASt3R-SLAM, VGGT-SLAM 2.0 and ViSTA-SLAM (columns).
  Dashed grey is ground truth. Colour shows per-frame ATE after alignment (blue $\to$ red, common scale capped at $100$\,mm). Each trajectory is aligned with Sim(3) on its first $40\,\%$ of poses and then left unaltered; Table~\ref{tab:results} uses SE(3) for ours.
  Each row uses the projection plane that the airway's ground truth moves in; a baseline whose estimate leaves the airway also leaves the panel.
  Numerical ATE values are in Table~\ref{tab:results}.}
  \label{fig:trajectories}
\end{figure*}

\subsection{Exp 2 -- Pose Tracking and Reconstruction Accuracy}
\label{sec:results}

\begin{figure*}[t]
  \centering
  \includegraphics[width=0.9\linewidth]{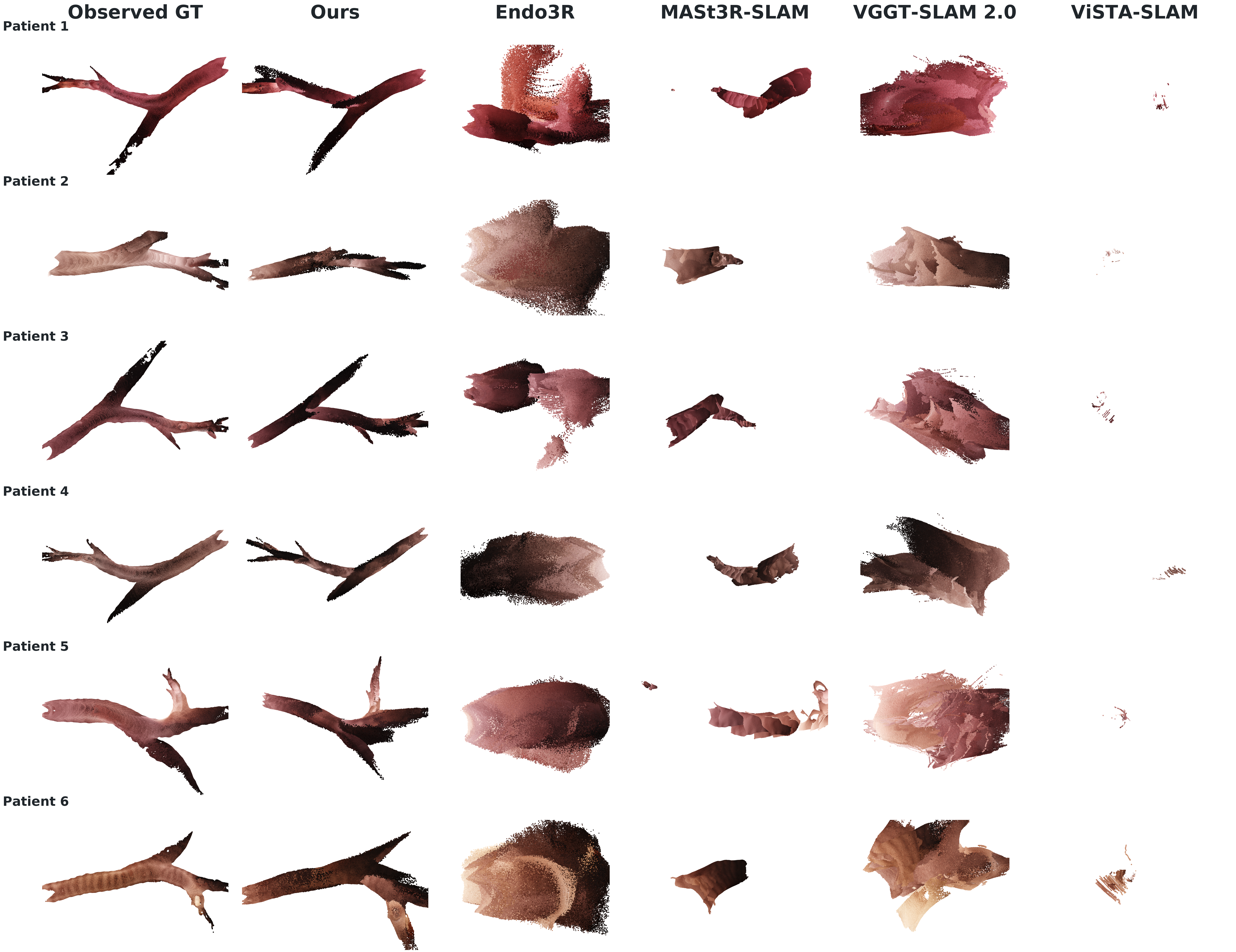}
  \caption{\textbf{Qualitative comparison of airway-surface reconstruction.} Ground truth (GT) surface points and reconstructions from our method and four dense SLAM baselines, on the six CT-derived airways (rows: synthetic airways rendered from the CT scans of Patients 1 to 6, all ATM'22 cases~\cite{zhang2023atm22}; Sec.~\ref{sec:setup}). We register predicted clouds to GT rigidly, and every method keeps the same projection and scale. We crop each panel to the ground truth depth range and mark a method with nothing left inside it accordingly. We omit ORB-SLAM3 for the reconstruction evaluation because of its sparse map.}
  \label{fig:reconstruction_results}
\end{figure*}

We next evaluate the accuracy of the estimated camera trajectories. Fig.~\ref{fig:trajectories} illustrates aligned trajectories and per-frame ATE, and Fig.~\ref{fig:reconstruction_results} shows the reconstructed surfaces beside the observed ground truth. Table~\ref{tab:results} reports ATE, F-score, CD and HD on all six airways.

ORB-SLAM3 completes only two of the six airways, at $20.84$ and $6.98$\,mm ATE. The smooth mucosal surface yields too few corner matches to keep successful tracking. We mark the failed sequences as $\times$. 
We do not report the surface metrics for the cases when the algorithms only output a sparse map (indicated by ``--'').

\begin{table*}[t]
\centering
\scriptsize
\renewcommand{\arraystretch}{1.0}
\caption{\textbf{Tracking and reconstruction metrics on the six CT-derived airways} (Patients 1 to 6, Sec.~\ref{sec:setup}). Our method is aligned with SE(3) and keeps its metric scale; every other method is aligned with Sim(3). The alignments are computed using the first $40\,\%$ of poses. F-score at $10$\,mm. ``$\times$'' marks tracking failure or under $80\,\%$ temporal coverage; ``--'' marks a sparse map, which has no surface to evaluate. $^{\dagger}$The mean is over the entries of the row that are not ``$\times$''.}
\label{tab:results}
\begin{tabular*}{0.8\textwidth}{c l c c c c c c c}
\toprule
& \textbf{Method} & \textbf{Patient 1} & \textbf{Patient 2} & \textbf{Patient 3} & \textbf{Patient 4} & \textbf{Patient 5} & \textbf{Patient 6} & \textbf{Mean}$^{\dagger}$ \\
\midrule
\multirow{6}{*}{\rotatebox[origin=c]{90}{ATE (mm) $\downarrow$}}
 & \textbf{Ours}                          & 26.42 & 34.19 & 28.26 & 34.53 & 22.57 & 10.83 & 26.13 \\
 & ORB-SLAM3~\cite{campos2021orbslam3}    & $\times$ & $\times$ & $\times$ & 20.84 & $\times$ & 6.98 & 13.91 \\
 & Endo3R~\cite{guo2025endo3r}            & 52.36 & 52.10 & 163.42 & 107.05 & 62.00 & 84.39 & 86.89 \\
 & MASt3R-SLAM~\cite{murai2025mast3rslam} & 15.17 & 12.35 & 23.62 & 23.07 & 29.12 & 28.77 & 22.02 \\
 & VGGT-SLAM 2.0~\cite{maggio2026vggtslam2} & 321.39 & 46.73 & 111.57 & 53.46 & 72.66 & 158.75 & 127.43 \\
 & ViSTA-SLAM~\cite{zhang2025vistaslam}   & 33.41 & 17.64 & 47.60 & 40.81 & 18.28 & 30.27 & 31.34 \\
\midrule
\multirow{6}{*}{\rotatebox[origin=c]{90}{F-score (\%) $\uparrow$}}
 & \textbf{Ours}                          & 75.54 & 79.76 & 70.38 & 77.05 & 79.17 & 82.94 & 77.47 \\
 & ORB-SLAM3                              & -- & -- & -- & -- & -- & -- & -- \\
 & Endo3R                                 & 66.98 & $\times$ & 1.31 & 2.04 & 31.17 & 17.05 & 23.71 \\
 & MASt3R-SLAM                            & 81.85 & 80.92 & 75.37 & 72.82 & 76.22 & 57.83 & 74.17 \\
 & VGGT-SLAM 2.0                          & $\times$ & 7.81 & 0.00 & 3.18 & 0.00 & $\times$ & 2.75 \\
 & ViSTA-SLAM                             & 30.12 & 38.12 & 20.06 & 21.76 & 30.78 & 16.95 & 26.30 \\
\midrule
\multirow{6}{*}{\rotatebox[origin=c]{90}{CD (mm) $\downarrow$}}
 & \textbf{Ours}                          & 9.18 & 7.65 & 13.92 & 9.05 & 8.28 & 7.72 & 9.30 \\
 & ORB-SLAM3                              & -- & -- & -- & -- & -- & -- & -- \\
 & Endo3R                                 & 13.71 & $\times$ & 32.28 & 37.63 & 17.85 & 24.83 & 25.26 \\
 & MASt3R-SLAM                            & 8.28 & 11.22 & 13.97 & 13.71 & 9.03 & 17.80 & 12.33 \\
 & VGGT-SLAM 2.0                          & $\times$ & 33.30 & 90.63 & 69.22 & 66.23 & $\times$ & 64.84 \\
 & ViSTA-SLAM                             & 28.57 & 19.31 & 34.09 & 34.57 & 21.53 & 27.64 & 27.62 \\
\midrule
\multirow{6}{*}{\rotatebox[origin=c]{90}{HD (mm) $\downarrow$}}
 & \textbf{Ours}                          & 39.68 & 28.50 & 60.86 & 48.39 & 35.33 & 41.74 & 42.42 \\
 & ORB-SLAM3                              & -- & -- & -- & -- & -- & -- & -- \\
 & Endo3R                                 & 67.92 & $\times$ & 77.65 & 117.36 & 49.46 & 84.02 & 79.28 \\
 & MASt3R-SLAM                            & 54.35 & 49.98 & 69.32 & 64.27 & 32.73 & 73.99 & 57.44 \\
 & VGGT-SLAM 2.0                          & $\times$ & 88.80 & 195.53 & 162.96 & 151.39 & $\times$ & 149.67 \\
 & ViSTA-SLAM                             & 102.53 & 66.71 & 120.32 & 118.01 & 75.94 & 88.00 & 95.25 \\
\bottomrule
\end{tabular*}
\end{table*}

Among the feed-forward systems, MASt3R-SLAM is the strongest baseline ($22.02$\,mm mean ATE, $74.17\,\%$ mean F-score). VGGT-SLAM 2.0 drifts the most ($127.43$\,mm mean ATE) and on two airways places no geometry inside the ground truth volume at all. Endo3R reaches $86.89$\,mm mean ATE despite endoscopic training data. ViSTA-SLAM accumulates less drift than either, yet fragments the surface (F-score $26.30\,\%$).

\emph{Ours} has the best mean F-score, Chamfer distance and Hausdorff distance of Table~\ref{tab:results}: $77.47\,\%$ against $74.17\,\%$ for MASt3R-SLAM, $9.30$\,mm against $12.33$\,mm, and $42.42$\,mm against $57.44$\,mm. Its mean ATE of $26.13$\,mm is higher than the $22.02$\,mm of MASt3R-SLAM, which is aligned with a free scale; ORB-SLAM3 has the lowest mean ($13.91$\,mm) but only on the two airways it completes. It is also the only one of the six systems that produces metric outputs.

Fig.~\ref{fig:reconstruction_results} shows the spatial structure behind these numbers. MASt3R-SLAM produces an accurate reconstruction but it covers only part of each airway. Endo3R and VGGT-SLAM 2.0 return clouds that extend well beyond the anatomy, and Endo3R on one airway and VGGT-SLAM 2.0 on two leave nothing inside the volume the ground truth occupies. ViSTA-SLAM fragments the surface, whereas our reconstructions preserve both the curvature of the lumen and the angle of bifurcations.

\subsection{Exp 3 -- Ablation Study}
\label{sec:ablation}

\begin{table}[!t]
\centering
\footnotesize
\renewcommand{\arraystretch}{1.0}
\setlength{\tabcolsep}{4pt}
\caption{\textbf{Ablation study} on the six synthetic phantom sequences (means over B00, B46, B71, B96, B97 and B118; Sec.~\ref{sec:setup}; not the CT-derived airways of Table~\ref{tab:results}), each row removing or replacing one component. ATE and CD in mm, F-score in \%; best per column in \textbf{bold}.
``$\times$'' indicates tracking failure.
}
\label{tab:ablation}
\begin{tabular*}{\columnwidth}{@{\extracolsep{\fill}} l c c c @{}}
\toprule
\textbf{Configuration} & \textbf{ATE}\,$\downarrow$ & \textbf{F-score}\,$\uparrow$ & \textbf{CD}\,$\downarrow$ \\
\midrule
\textbf{Ours (full)}                 & \textbf{6.12} & \textbf{95.3} & \textbf{4.10} \\
\;\;ORB instead of DINO matcher      & $\times$ & $\times$ & $\times$ \\
\;\;w/o TSDF map and ICP factor & 16.34 & 64.9 & 13.97 \\
\;\;single-pass fusion               & 18.88 & 70.2 & 15.95 \\
\;\;w/o GNC (initial scale)           & 27.47 & 55.3 & 23.55 \\
\bottomrule
\end{tabular*}
\end{table}

We remove one component at a time and rerun the six phantom sequences. Table~\ref{tab:ablation} reports mean ATE, F-score and Chamfer distance for the following variants.
\begin{itemize}[nosep,leftmargin=*]
\item \emph{ORB instead of DINO matcher:} we match ORB features in place of the DINOv3 descriptors of Sec.~\ref{sec:image_to_map}.
\item \emph{w/o TSDF map and ICP factor:} instead of the ICP factor of Sec.~\ref{sec:icp}, the graph contains the visual factors only.
\item \emph{Single-pass fusion:} instead of optimising the graph again after adding the ICP factor, we add it and optimise at the next keyframe.
\item \emph{w/o GNC:} instead of the schedule of~\eqref{eq:welsch}, ICP uses the initial Welsch scale $\tau_0$ throughout.
\end{itemize}
ORB breaks tracking: its matches on the smooth mucosa are too few, and every phantom halts at the first unstable keyframe. Among the ICP variants, removing GNC degrades the system the most, raising ATE from $6.1$ to $27.5$\,mm and lowering F-score from $95.3$ to $55.3\,\%$. Without GNC, ICP converges in $30\,\%$ of attempts instead of $94\,\%$, and only $355$ factors are accepted against $1{,}094$. Single-pass fusion integrates each keyframe into the TSDF at its uncorrected pose. It reaches $18.9$\,mm while accepting almost as many ICP factors ($1{,}063$). Removing the TSDF map and ICP factor gives $16.3$\,mm and $64.9\,\%$. On B97 and B118, the two shortest branches, every ICP variant reaches at most $3.6$\,mm ATE; the means are driven by the other four.

\subsection{Exp 4 -- Validation on Real Bronchoscopy}
\label{sec:real_bronch}

We first compare against the baselines on two held-out ex vivo sequences. Neither has a ground truth pose, so we report no ATE or F-score and inspect them qualitatively. We then evaluate the trajectory on the ex vivo sequences that carry EMT ground truth. Beyond the ventilation already noted, the ex vivo sequences add moving specular highlights from the co-located illumination, motion blur, appearance differences between specimens, and mucus and vasculature absent from the renderings.

\begin{figure*}[t]
  \centering
  \includegraphics[width=\linewidth]{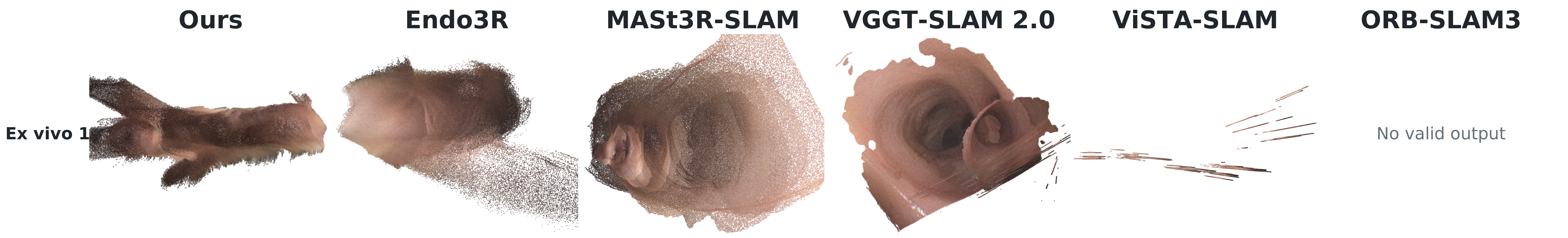}
  \caption{\textbf{Reconstruction comparison on ex vivo tissue.} Each panel is framed on its own extent because real sequences carry no ground truth pose and monocular scale is free. The comparison therefore judges structural coherence. An empty panel is itself the result.}
  \label{fig:real_baseline_comparison}
\end{figure*}

No baseline recovers a usable endoluminal reconstruction on these sequences, as Fig.~\ref{fig:real_baseline_comparison} shows. ORB-SLAM3 does not initialise on the low-texture mucosa: it retains no map point and writes an empty trajectory. MASt3R-SLAM and VGGT-SLAM 2.0 return dense but degenerate geometry, with the lumen collapsed into a single frontal sheet. ViSTA-SLAM fragments the airway into disconnected pieces, and its own confidence threshold rejects every pixel of the fused cloud. The panel therefore shows its output with that threshold relaxed to keep the most confident half of its predictions. Endo3R recovers a tubular structure but leaves a diffuse cloud around it. Of the methods compared, only ours returns a coherent lumen that keeps its diameter.

\begin{figure}[!t]
  \centering
  \includegraphics[width=\columnwidth]{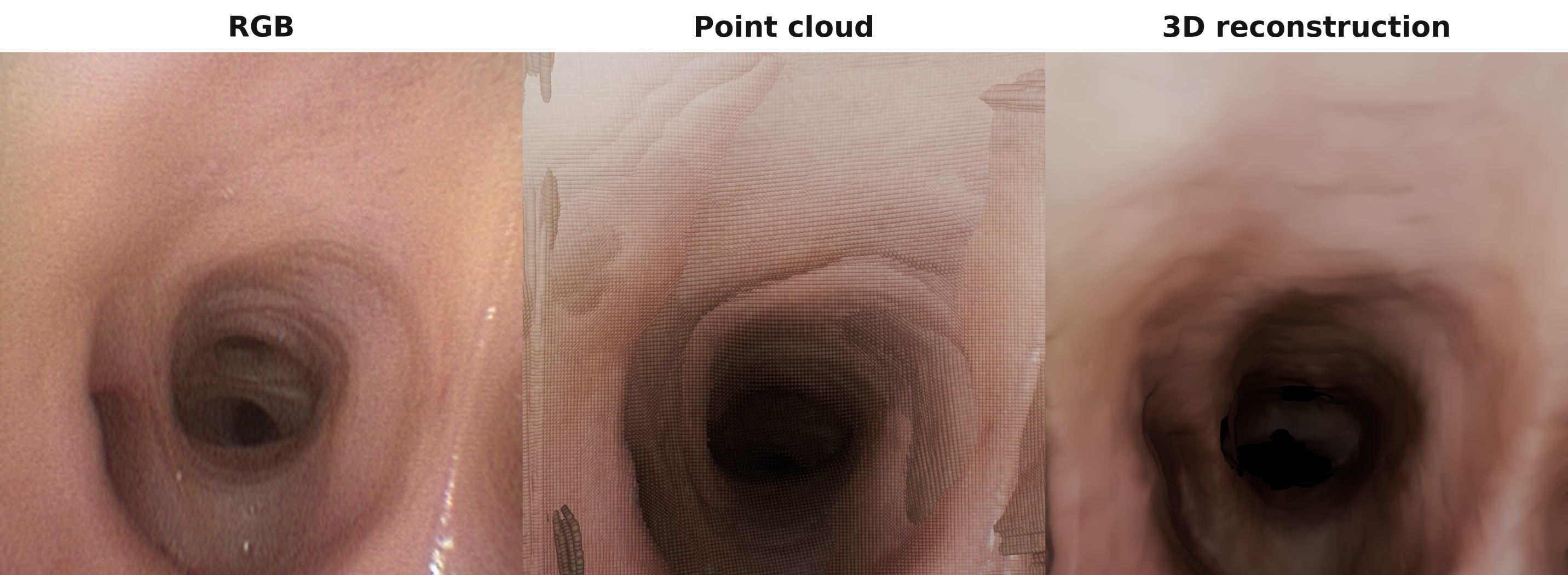}
  \caption{\textbf{From bronchoscope frame to endoluminal surface.} Left: input RGB frame from an ex vivo bronchoscopy sequence. Centre: dense point cloud obtained by back-projecting the DA$^3$ depth at the SLAM-estimated pose. Right: corresponding region of the fused TSDF surface after marching-cubes extraction and light smoothing.}
  \label{fig:comparison_subfigure}
\end{figure}

Fig.~\ref{fig:comparison_subfigure} shows the reconstructions from the camera's viewpoint. The per-keyframe point cloud (centre panel) reproduces the lumen geometry visible in the input RGB frame, and the fused TSDF mesh (right panel) joins those local observations into one continuous surface.

\begin{figure*}[t]
  \centering
  \includegraphics[width=1\linewidth]{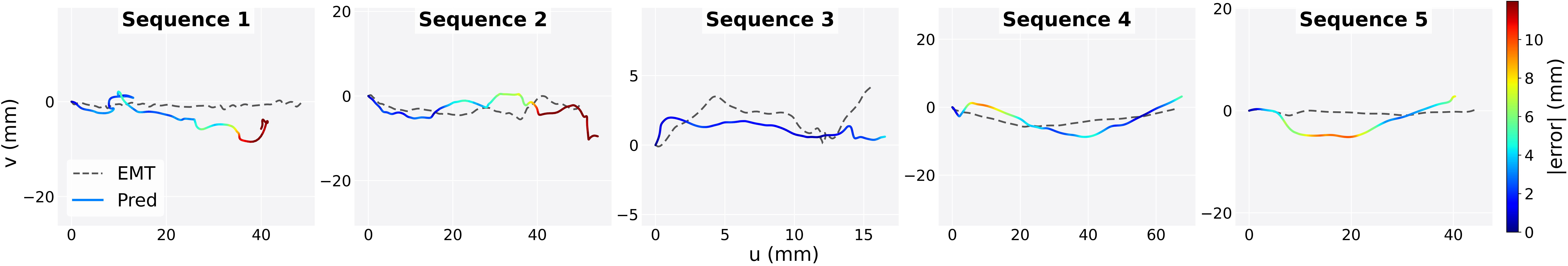}
  \caption{\textbf{Estimated trajectory against electromagnetic-tracking ground truth, evaluated as odometry.} Five acquisition windows, each projected onto its dominant plane $(u,v)$. Dashed grey is the EMT reference over its longest continuous valid stretch. We align the estimate with a similarity transform fitted on the start of each window (Sec.~\ref{sec:real_bronch}), and the colour (shared bar, mm) shows the drift accumulated since the start.}
  \label{fig:emt_trajectories}
\end{figure*}

\begin{figure*}[t]
  \centering
  \includegraphics[width=\linewidth]{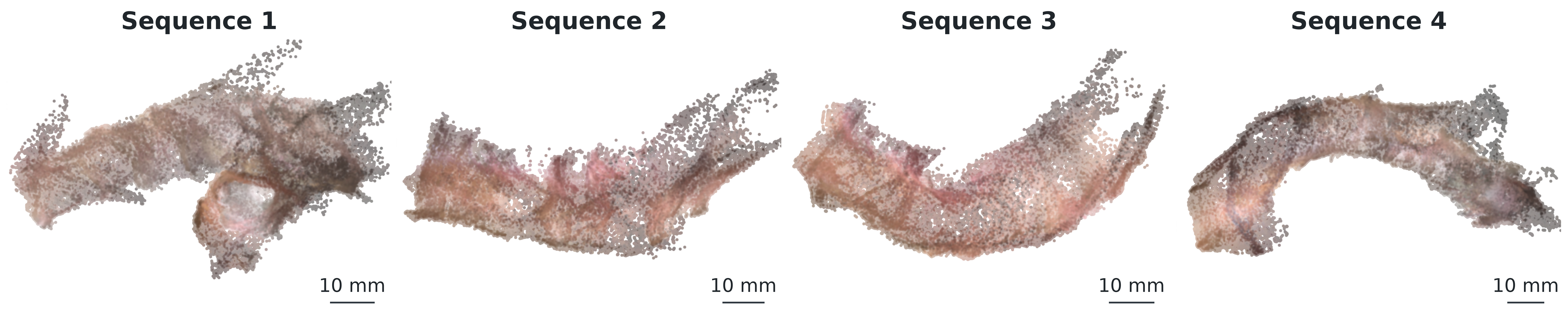}
  \caption{\textbf{Endoluminal reconstruction on ex vivo tissue.} Graph-optimised sparse landmark maps for four of the five EMT windows of Fig.~\ref{fig:emt_trajectories}, metrically scaled and rendered as depth-ordered surfels; colour is the observed mucosal appearance and the scale bar is $10$\,mm. The maps recover a continuous lumen with visible cartilage ridging and bifurcation openings.}
  \label{fig:emt_landmark_clouds}
\end{figure*}

\vspace{0.5em}
\noindent\textbf{Comparison against electromagnetic tracking.} On the ex vivo sequences with EMT ground truth, we evaluate the windows in which the scope travels at least $54$\,mm in the anatomy frame, measured on the EMT reference.

Because these traversals contain no revisits, we evaluate them as \emph{odometry}. We fit a similarity transform on the initial portion of each window and then apply it unchanged, so downstream drift stays visible. Fig.~\ref{fig:emt_trajectories} shows five windows under this protocol, coloured by per-sample deviation. Fig.~\ref{fig:emt_landmark_clouds} shows the corresponding reconstructions on real tissue, to compare with the synthetic surfaces of Fig.~\ref{fig:reconstruction_results}.

The electromagnetic sensor and the camera sit on either side of the actuated bending section, which gives them a non-rigid offset, and the reference is the sensor origin. The specimen is also mechanically ventilated, and the reference contains a respiratory oscillation that the camera, which moves with the tissue, does not observe. We band-stop the measured ventilation frequency before comparison.

\subsection{Exp 5 -- Photorealistic Rendering}
\label{sec:gaussian}

The optimised poses also support appearance reconstruction: we train a 3D Gaussian Splatting model~\cite{kerbl2023gaussian} offline on the four EMT sequences, seeding Gaussians from the depth-backprojected keyframe cloud with per-segment SE(3) pose refinement. Rendering held-out views tests whether the estimated poses are mutually consistent.

\begin{figure}[!t]
  \centering
  \includegraphics[width=\columnwidth]{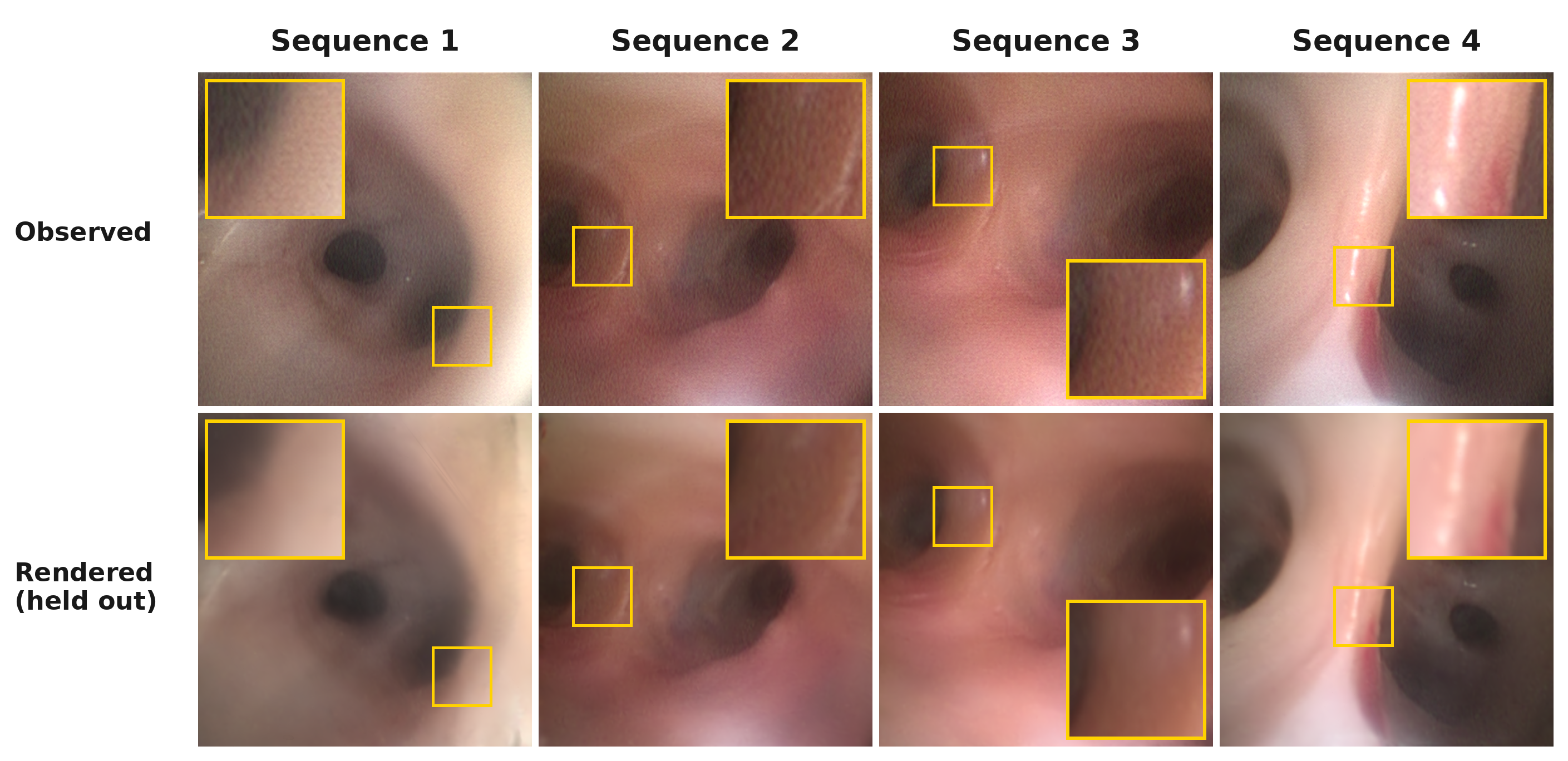}
  \caption{\textbf{Novel-view synthesis from the estimated trajectory.} For each of the four sequences, an observed frame above the rendering of that same held-out view; the yellow box marks a detailed region, magnified in the inset. See the supplementary video for a closer look.}
  \label{fig:gaussian_showcase}
\end{figure}

\begin{table}[!t]
\centering
\footnotesize
\renewcommand{\arraystretch}{1.0}
\setlength{\tabcolsep}{6pt}
\caption{\textbf{Novel-view synthesis, per sequence.} A $3$D Gaussian Splatting model trained from the estimated poses of each EMT sequence. Held-out views are excluded from training on a global index.
}
\label{tab:gaussian}
\begin{tabular}{l c c c c}
\toprule
\textbf{Sequence} & \textbf{Views} & \textbf{Segments} & \textbf{PSNR train} & \textbf{PSNR held out} \\
\midrule
1 & $647$ & $21$ & $35.18$ & $33.86$ \\
2 & $270$ & $9$  & $35.33$ & $34.70$ \\
3 & $270$ & $9$  & $35.43$ & $34.53$ \\
4 & $337$ & $11$ & $35.55$ & $33.98$ \\
\midrule
\textbf{Mean} & & & $\mathbf{35.37}$ & $\mathbf{34.27}$ \\
\bottomrule
\end{tabular}
\end{table}

Every eighth view is held out. Fig.~\ref{fig:gaussian_showcase} shows observed frames beside renderings of the same held-out views. Held-out PSNR averages $34.3$\,dB against $35.4$\,dB on training views (Table~\ref{tab:gaussian}), within $0.5$\,dB of that mean on every sequence, for sequences of $270$--$647$ keyframes.

\subsection{Computational Performance}
\label{sec:compute}

We report per-keyframe timings on a single NVIDIA RTX 4090, averaged over the $1{,}132$ keyframes of the six CT-derived airway runs of Exp.~2, which select every third frame of $30$\,fps video. A keyframe is processed synchronously: its pose is published once tracking, registration and the graph update have finished. Tracking takes $277$\,ms (depth preprocessing $61$\,ms, matching $60$\,ms, landmark association $80$\,ms, PnP $76$\,ms), ICP against the local submap $168$\,ms, and the graph update with landmark birth $263$\,ms, of which both iSAM2 passes take $200$\,ms. Each keyframe is also processed on a copy of the map state, which replaces the committed state only after it is validated; the copy takes $1.86$\,s. A keyframe takes $2.98$\,s in total (median $2.87$\,s, 95th percentile $5.38$\,s), so the system runs slower than real time on these sequences. DINOv3 descriptors are read from a precomputed cache in these runs; extracting them online adds $200$\,ms per frame.

\subsection{Limitations}
\label{sec:limitations}

Heavily specular or textureless mucosa weakens both front-end signals at once. DA$^3$ flags most pixels as unreliable, which leaves ICP with too few source points against the local submap, and the DINO matcher finds too few correspondences to support the projection factors. ICP is also weakest over the first few keyframes, before the submap holds enough surface. Diagnosing the ICP Hessian lets the system discard a degenerate registration, and the visual factors alone then constrain the pose.

The system performs no loop closure, and each evaluated sequence traverses a single branch without revisiting it. Exploratory bronchoscopy that withdraws and re-enters a different branch would revisit mapped geometry, and closing those loops is required before the system can cover a full airway tree in one session.

On the ex vivo sequences no ground truth depth is available, so we can check the inherited metric scale only for internal consistency. We reconstruct under a rigid-scene assumption, which holds exactly for the static synthetic sequences. The ex vivo lungs are ventilated and breathe during acquisition, but the video shows little deformation of the airways, so we do not model it. We target the proximal-to-subsegmental airways, from the primary bronchi down to the $2$--$4$\,mm subsegmental branches, and have not tested larger tidal volumes or the distal airways, where deformation is likely larger. Modelling the deformation~\cite{lamarca2021defslam, gomezrodriguez2024nrslam, song2018misslam}, or reconstructing at a fixed phase of the breathing cycle, would address this.

\section{Conclusion}
\label{sec:conclusion}

We introduced a monocular SLAM system that recovers both the bronchoscope trajectory and a dense reconstruction of the inner airway surface, without depth sensors or preoperative scans. It combines a bronchoscopy-tuned depth predictor for metric scale, a dense matcher for texture-poor mucosa, and a two-pass factor graph that fuses those visual measurements with a dense ICP constraint against the system's own reconstruction.

On six CT-derived airways, ours has the most accurate surface of the six systems, with $77.47\,\%$ mean F-score against $74.17\,\%$ for the strongest baseline, and a mean ATE of $26.13$\,mm against $22.02$\,mm for MASt3R-SLAM. 

Our results demonstrate that accurate camera pose and a dense endoluminal surface can therefore be recovered from bronchoscopy video alone, with no added hardware and no preoperative volumetric prior.

\appendices
\section{}
\subsection{Metric Depth Predictor Training}
\label{app:depth_training}
\noindent\textbf{Architecture and parameterisation.} We fine-tune the nested DA$^3$ variant, which pairs the pretrained any-view geometry branch (\texttt{DA3NESTED-GIANT-LARGE-1.1}) with a metric dense prediction transformer (DPT) head. The metric head regresses depth in the canonical form $\mathbf{D}_c = M\,\mathbf{D}\,f^c\!/f$, with canonical focal $f^c = 300$\,px and a fixed multiplier $M = 100$; deployment inverts this, $\mathbf{D} = \mathbf{D}_c\,f/(f^c M)$, so predictions carry absolute millimetre scale under the target camera's calibrated intrinsics. Depths are supervised up to $200$\,mm, the working range of the bronchoscope.

\vspace{0.5em}
\noindent\textbf{Data and supervision.} Training uses over one million photorealistic airway renderings with per-pixel ground truth depth, sampled as clips of $4$ frames at stride $6$ so the loss sees the temporal context the deployed predictor is given. Supervision is metric: an $L_1$ term on canonical depth against raw ground truth, combined with a ray-direction term inherited from the any-view branch that regularises the implied camera geometry.

\vspace{0.5em}
\noindent\textbf{Optimisation.} AdamW, weight decay $0.05$, with separate rates for the pretrained backbone ($5\times10^{-6}$) and the metric head ($5\times10^{-5}$), $8000$ warm-up steps, and one clip per step with gradient accumulation over $8$, for $4.1\times10^{4}$ steps. Colour jitter on $80\,\%$ of samples and spatially correlated sensor noise keep the predictor from tuning to the renderer's noise-free statistics.

\vspace{0.5em}
\noindent\textbf{Uncertainty channel.} The metric head emits a second channel alongside depth, carrying a Laplace scale in \emph{log}-depth. Its activation is $\exp(\cdot)+1$, so the log-scale parameter is recovered exactly as $\log b = \log(c - 1)$ from the emitted value $c$, clamped to $[-12,4]$ and multiplied by a metric temperature fitted once at calibration. Because the Laplace lives in log-depth, the $95\,\%$ interval on depth is $\mathbf{D}\exp(\pm q_{95} b)$ with $q_{95} = -\ln 0.05$, and we report the standard deviation $\mathbf{S}_t$ as the Gaussian-equivalent width of that interval,
\begin{equation}
\mathbf{S}_t = \mathbf{D}\,\frac{e^{\,q_{95} b} - e^{-q_{95} b}}{2 \times 1.960}.
\label{eq:equiv_std}
\end{equation}
Two auxiliary heads supply the per-frame validity mask, one predicting whether a pixel falls inside the trained depth range and one rejecting background; pixels failing either are discarded before~\eqref{eq:reliability} is evaluated.

\subsection{Cross-Attention Matcher Training}
\label{app:matcher_training}
The DINO feature matcher of Sec.~\ref{sec:image_to_map} maps the DINOv3 patch descriptors of a source and a target frame into a $256$-D matching space and predicts coarse patch assignments and a sub-pixel offset, supervised on frame pairs with known correspondences.

\vspace{0.5em}
\noindent\textbf{Supervision.} Ground truth correspondences come from frame pairs with known dense flow: pairs are precomputed once, each patch of the source frame carrying either a target location, an explicit invalid label where the surface leaves the field of view, or an ignore label where the correspondence is occluded or uncertain. The third state matters: the mucosa self-occludes at every fold, and a two-state formulation would train the matcher to guess there. The loss excludes occluded patches.

\vspace{0.5em}
\noindent\textbf{Losses.} We consider a cross-entropy over coarse patch assignment; a Charbonnier regression on the sub-pixel offset, evaluated on valid patches only; and a forward-backward cycle-consistency term (weight $0.5$) that penalises correspondences which do not return to their origin. These terms are applied in both directions for each pair. 

\vspace{0.5em}
\noindent\textbf{Optimisation.} Descriptors are precomputed with the frozen backbone, which makes large batches affordable: AdamW, batch size $64$, learning rate $3\times10^{-4}$, weight decay $0.01$, one warm-up epoch, on $400$\,px crops. Training runs to a maximum of $20$ epochs with early stopping (patience $4$ epochs, minimum improvement $0.5\,\%$); the deployed checkpoint is the best of $15$ completed epochs.

\section*{Acknowledgments}
This work was supported by a UKRI Turing AI World Leading Researcher Fellowship on AI for Person-Centred and Teachable Autonomy (grant EP/Z534833/1) and Baillie Gifford Pandemic Science Hub.

\balance
\bibliographystyle{IEEEtran}
\bibliography{references}

\end{document}